\documentclass{article}
\usepackage{graphicx}
\usepackage[style=numeric, citestyle=numeric,sorting=none]{biblatex}
\usepackage{amsfonts, amsmath, amsthm, amssymb}
\usepackage{booktabs}
\usepackage{float}
\usepackage[left=1in,top=1in,right=1in,bottom=1in]{geometry}
\usepackage[hidelinks]{hyperref}
\usepackage{setspace}
\newcommand{\RR}[0]{\mathbb{R}}

\newcommand{\Var}[0]{\text{Var}}

\title{A comparison of knot point screening algorithms for highly adaptive lasso}
\author{Yadi Weng$^1$, Sky Qiu$^2$\\
University of North Carolina, Chapel Hill$^1$\\
Division of Biostatistics, University of California, Berkeley$^2$\\
{\tt yadiw@unc.edu}}
\date{\today}

\begin{document}
\maketitle
\begin{abstract}
Highly adaptive lasso (HAL) is a nonparametric regression estimator with a rate of convergence $n^{-1/3}(\log n)^{d/2}$, where the data dimension $d$ only appears in the power of the log of sample size. This fast rate is crucial for many causal inference tasks, where efficient estimation typically requires nuisance parameters to be estimated at a specific rate. HAL achieves this rate and, when paired with semiparametrically efficient estimators—such as those constructed using targeted maximum likelihood estimation (TMLE)—can be used to develop efficient estimators. Beyond causal inference, HAL is applicable to a wide range of supervised machine learning tasks. The procedure for HAL involves first performing a basis expansion on the original design matrix $X$ and then computing a lasso solution path. Cross-validation is used to select the optimal $L_1$-norm for the coefficients. For datasets with all continuous covariates and no ties, the HAL design matrix can reach a dimension as large as $n\times n(2^d-1)$, imposing a significant computational and memory burden. To address this, screening algorithms are applied at both the variable and knot-point levels to make HAL practical for datasets of moderate to large sizes. In this work, we propose outcome-blind knot point screening based on center-based clustering---$k$-means and partitioning around medoids (PAM)---and compare it with the quantile-based screening implemented in the \texttt{hal9001} R package at matched design-matrix sizes. Across 153 simulation settings, $k$-means and PAM significantly improved test $R^2$ in 97 and 111 settings, respectively, with gains that grow with covariate skewness, while producing slightly smaller design matrices. On nine real-world datasets, they reduced the cross-validated mean squared error by 35--36\% on the dataset with the most skewed predictors and performed comparably on the others.
\end{abstract}

\section{Introduction}
Regression is a fundamental statistical estimation problem which involves identifying a function mapping from independent variables $X\in\RR^d$ to a dependent variable $Y\in\RR$ that minimizes the empirical mean of some loss function \cite{witten_islr_2013}. Regression problems are common in the field of causal inference. For example, one might be interested in estimating the average treatment effect (ATE) based on data generated from either interventional or non-interventional studies. A good estimator for ATE ought to be efficient, formally defined as no other estimators with a smaller sampling variance exist \cite{casella_berger}. To construct an efficient ATE estimator, such as one developed through targeted maximum likelihood estimation (TMLE) \cite{vdl_tmle_2006,tmle_book_2011}, which updates an initial estimator with a parametric submodel through it to achieve an optimal bias-variance trade-off for the target parameter, it is necessary to estimate both the conditional mean of outcome and the conditional probability of treatment, two regression problems. To guarantee TMLE's efficiency for the ATE target parameter, both nuisance estimators need to converge their true values at a rate faster than $n^{-1/4}$ \cite{tmle_book_2011,tsiatis_semi_2006}. Parametric maximum likelihood estimators (MLE) such as the least-squares linear regression satisfy the rate condition. However, they often operate under unrealistic assumptions on the model class containing the true function. As in linear regression, one assumes that the true function is a linear combination of $X$, a restriction that is unlikely to hold true in many practical settings. On the other hand, nonparametric regression methods such as kernel regression \cite{nadaraya_regression_1964,watson_regression_1964} offers a flexible alternative but fail to achieve a fast-enough convergence rate unless local-smoothness assumptions are imposed. In recent years, highly adaptive lasso (HAL) has emerged as the sole nonparametric regression estimator capable of achieving fast convergence rates without imposing stringent local-smoothness assumptions on the function class \cite{hal_2016}. HAL has a convergence rate of $n^{-1/3}(\log n)^{d/2}$, with this rate depending only on the dimension $d$ via a power of the $\log n$-factor, hence essentially dimension-free \cite{bibaut_fast_2019}. HAL only assumes a mild assumption that the true function is \textit{càdlàg} (right-continuous with left limits) with a bounded variation norm which includes most functions one would encounter in fields like public health and medicine.

Despite these notable theoretical advantages, HAL's practical application is limited by its large memory requirement computationally, rendering it infeasible to run on moderate to large-sized datasets \cite{schuler_sal_2022}. The current HAL implementation (R package \texttt{hal9001} \cite{hal9001_cran,hal9001_paper}) recognizes HAL's equivalence in terms of the optimization objective to that of lasso (least absolute shrinkage and selection operator) \cite{tibshirani_lasso_1996}, utilizing lasso software packages such as \texttt{glmnet} in R for optimization. However, in order to use lasso software to run HAL, one needs to first augment the data and in the worst-case scenario, where all $d$ variables are continuous without ties, the HAL augmented design matrix has a dimension of $n$ by $n(2^d-1)$. This results in considerable computational strain. Addressing these computational challenges requires significant storage capacity that far exceeds the capabilities of standard personal computing devices, even when leveraging more efficient storage options like sparse matrices. In this work, we discuss several existing and new methods aiming at addressing the memory issue of HAL. We evaluate the performance of these methods on both simulated and real-world datasets, and find that clustering-based knot screening matches or improves upon the quantile-based default in most settings while producing slightly smaller design matrices, with the largest gains when the covariates are skewed.

The remainder of this paper is organized as follows. In Section \ref{sec:hal}, we provide a gentle introduction to HAL, its theoretical properties, and its computational bottlenecks. Section \ref{sec:framework} outlines a modular screening framework for controlling memory usage. Section \ref{sec:screening} introduces the three outcome-blind knot point screening algorithms—quantile-based discretization, $k$-means clustering, and partitioning around medoids (PAM)—along with our basis-budget matching scheme. Sections \ref{sec:sim} and \ref{sec:real} present empirical evaluations across 17 simulation designs and nine real-world datasets, respectively. Section \ref{sec:discussion} discusses the findings, practical recommendations, and directions for future work.

\section{Highly adaptive lasso}\label{sec:hal}
Define a random variable $O=(X,Y)$, where $X\in\RR^d$ and $Y\in\RR$ denote the independent and dependent variables, respectively. We observe $n$ independent and identically distributed (i.i.d.) copies of $O$, denoted by $O_1, O_2, \dots, O_n$, drawn from the true data-generating distribution $P_0$. We use $P_n$ to denote the empirical distribution of $O.$ Let $Pf=\int f(o)dP(o)$ and $P_nf=1/n\sum_{i=1}^nf(o_i)$ for a function $f(o).$ The highly adaptive lasso (HAL) estimator \cite{hal_2016} minimizes the empirical mean of a loss function over a specific function class of $d$-dimensional real-valued \textit{càdlàg} (right-continuous with left limits) functions \cite{neuhaus_cadlag}. The function class HAL minimizes over, denoted by $\mathcal{F}_M([0,1]^d)$, consists of functions with a bounded sectional variation norm, $M$. We define the sectional variation norm, $||f||_v$, of a function $f$ as:
$$
||f||_v=|f(0)|+\sum_{s\subset\{1,\dots,d\}}\int|df(x_s)|,
$$
where $f(x_s)$ denotes the function $f$ where the coordinates not in $s$ are set to 0. This norm (a.k.a.\ uniform sectional variation norm \cite{gill_inefficient_1995} or Hardy-Krause variation \cite{fang_multivariate_2021}) essentially measures the total amount of variation of a function across all coordinates of its domain. The HAL estimator, $f_n$, is then given by:
$$
f_n=\arg\min_{f\in \mathcal{F}_M([0,1]^d)} P_nL(f),
$$
where $L(f)$ is a loss function, such as the squared-error loss. In particular, we can approximate any function $f\in\mathcal{F}_M([0,1]^d)$ by
$$
f_m(x)=f(0)+\sum_{s\subset\{1,2,\dots,d\}}\sum_{j}I(x_s\geq u_{s,j})df_{m,s,j}.
$$
This approximation is obtained by applying a discrete measure to approximate the sectional representation of \textit{càdlàg} functions \cite{gill_inefficient_1995,hal_2016}. Intuitively, this approximation could be viewed as a linear combination of basis functions $\{\phi_{s,j}(x)=I(x_s\geq u_{s,j}):s,j\}$ with coefficients $\{df_{m,s,j}:s,j\}$. Also, notice that sectional variation norm corresponds exactly to the absolute sum of the coefficients and $|f(0)|$. Specifically, we have
$$
||f_m||_v=|f(0)|+\sum_{s\subset\{1,2,\dots,d\}}\sum_{j}|df_{m,s,j}|\leq M.
$$
Now, the problem of trying to find an empirical risk minimizer in the \textit{càdlàg} function class with bounded sectional variation norm becomes finding a linear combination of basis functions whose sum of absolute values of the coefficients including the intercept are bounded by a constant $M$. This is equivalent to a lasso optimization problem with the constraint that the $L_1$-norm of the coefficients are bounded by $M$.

Once the connection to lasso is made, then the implementation of HAL becomes clear. First, we apply basis expansion to the original design matrix $X$ to obtain a HAL design matrix. Then, we run lasso on the HAL design matrix, using cross-validation to select the penalization factor, which corresponds with an $L_1$-norm value. In this note, we focus on a sufficient initial set of basis functions directly implied by the data itself following \cite{hal_2016}.

\subsection{Constructing the HAL design matrix}
To construct the zero-order HAL design matrix, each basis function $\phi_{s,i}$ is defined as an indicator function $\phi_{s,i}(X) = I(X_s \geq x_{s,i})$, where $x_{s,i}$ denotes the $i$-th observation restricted to the covariate index subset $s \subset \{1,2,\dots,d\}$. Specifically, we begin with one-way basis functions: for each covariate index $j \in \{1,2,\dots,d\}$, we construct up to $n$ indicator basis functions $\phi_{\{j\},i}(X) = I(X_j \geq x_{\{j\},i})$, where $x_{\{j\},i}$ is the $i$-th realization of $X_j$. Two-way basis functions are then formed as products of one-way indicator basis functions evaluated at observed pairs, and higher-order basis functions are constructed analogously up to $d$-way interactions. For example, with $d=2$ variables and $n=2$ observations $X_1 = (2.1, 4.3)$ and $X_2 = (1.2, 5.6)$, this construction yields 6 basis functions: $I(X_1 \geq 2.1)$, $I(X_1 \geq 4.3)$, $I(X_2 \geq 1.2)$, $I(X_2 \geq 5.6)$, $I(X_1 \geq 2.1, X_2 \geq 1.2)$, and $I(X_1 \geq 4.3, X_2 \geq 5.6)$.

\subsection{Summary of HAL's theoretical properties}
HAL is noteworthy for its fast convergence in loss-based dissimilarity, with dimension $d$ depending only through the power of a $\log n$-factor. This is due to the fact that it falls in a function class that is a uniform Donsker with a well-behaved entropy integral \cite{vdl_efficient_2023}. Because of this, HAL avoids the curse of dimensionality to the extent that the dimension's influence on the convergence rate manifests solely through the logarithmic factor's exponent.

\subsection{Computational challenges of HAL}
\begin{figure}[ht]
\centering
\includegraphics[width=0.9\textwidth]{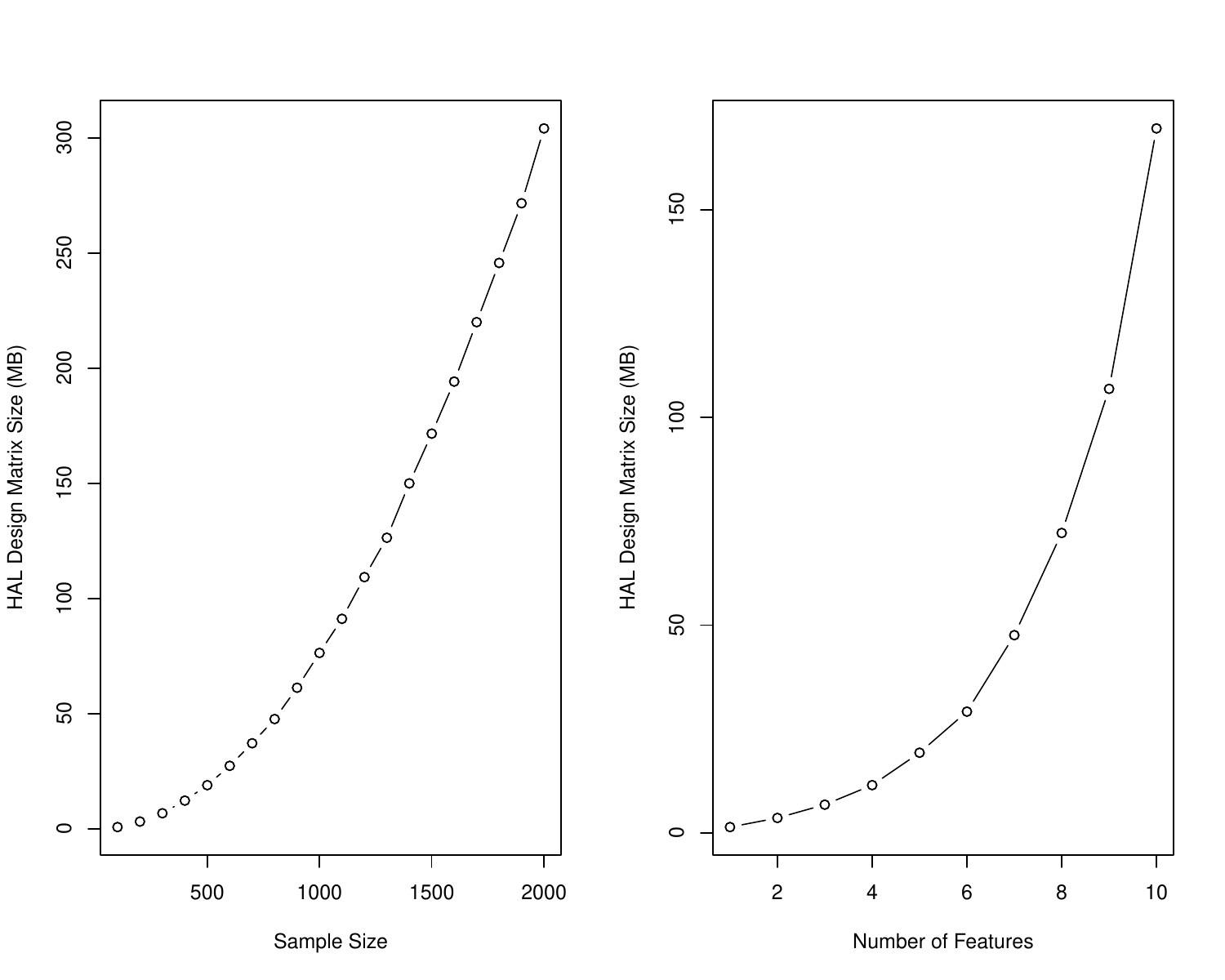}
\caption{Memory usage of the sparse HAL design matrix constructed by \texttt{hal9001} as a function of sample size $n$ and covariate dimension $d$.}
\label{fig:memory}
\end{figure}

The current implementation of the HAL estimator (R package \texttt{hal9001} \cite{hal9001_cran,hal9001_paper}) first creates the augmented HAL design matrix and then calls the \texttt{glmnet} package which runs the coordinate descent algorithm for optimization. The HAL design matrix has a dimensionality of at most $n \times n(2^d-1)$. This can rapidly exceed the memory capacity of standard personal computers, even when $n$ and $d$ are of moderate sizes. Figure \ref{fig:memory} shows that the memory of the HAL design matrix (stored in sparse matrix form) grows very fast as sample size and the number of features in the data grow.
Very quickly this would run out of memory, not to mention the optimization of \texttt{glmnet} on this huge matrix. Although this initial set of knot points is sufficient for HAL to attain its theoretical convergence rate, this set is not necessary. In particular, columns that are nearly constant (consisting almost entirely of ones or zeros) are unlikely to receive nonzero coefficients in \texttt{glmnet} and can be discarded beforehand. In the next section, we discuss a flexible modular framework that allows users to apply screening algorithms to reduce the size of the design matrix, thereby controlling memory usage and improving the speed of \texttt{glmnet}.

\section{A Modular Framework to Control Memory}\label{sec:framework}
As we have seen in the previous section, the key to scaling up HAL in high-dimensional data settings is to reduce the number of columns of the HAL design matrix. To achieve this goal, one can apply screening algorithms at three different levels. First, if there are hundreds of variables in the data, one should perform variable-level screening, such as running a main-term-only lasso and selecting the variables with nonzero coefficients, or applying another variable screening method. Second, one can perform screening at the basis-function level. Finally, for each basis function, one needs to screen the knot points. This screening framework is modular in the sense that a user can replace the screening algorithm at any level with their preferred choice.

\section{Knot point screening algorithms}\label{sec:screening}

\subsection{Quantile-based discretization}
We use the quantile-based knot selection procedure implemented in the \texttt{hal9001} R package \cite{hal9001_paper} as our reference method. To reduce the number of basis functions, the procedure discretizes each predictor using a specified number of knot points derived from its empirical distribution. The number of knots can vary by interaction degree, allowing higher-order interactions to use a smaller knot budget and limiting the combinatorial growth of the HAL design matrix. Specifically, let $B_1$, $B_2$, and $B_3$ denote the numbers of knot points for first-, second-, and third-degree basis functions. For a subset $s$ of $|s|$ predictors, each predictor in $s$ is discretized to $B_{|s|}$ of its empirical quantiles, and the distinct discretized observations are used as knot points. The knots are therefore spread evenly in probability, and a binary predictor contributes at most two distinct values.

For the simulation studies, we specify knot budgets of $(50, 15, 5)$ for first-, second-, and third-degree basis functions, respectively. For the real-world datasets, we consider first-degree knot budgets ranging from $20$ to $100$ in increments of $10$, while keeping the second- and third-degree budgets at $10$ and $5$. Knot selection is performed using the training covariates without reference to the outcome.

\subsection{$k$-means clustering-based quantizer}
Quantile-based discretization relies on univariate marginal quantiles and does not directly exploit the joint spatial geometry of the covariates. To address this challenge in moderate- to high-dimensional settings, we propose the derivation of a smaller knot-point set, $\Theta=\{\theta_{s,j}:s,j\}$, from the full data-implied set $\Phi=\{\phi_{s,j}:s,j\}$, where $|\Theta|<|\Phi|$. The smaller knot-point set $\Theta$ implies a smaller number of basis functions and thus a thinner HAL design matrix, thereby adhering to computational and memory limitations. Importantly, we remark that $\Theta$ is not necessarily a subset of $\Phi$, depending on the clustering technique used. In this section, we discuss strategies for identifying an optimal knot-point set $\Theta$ through center-based clustering algorithms, notably the $k$-means algorithm. This method partitions $n$ observations into $k$ clusters, optimizing the selection of cluster centers $\theta_1,\dots,\theta_k$ to minimize the within-cluster sum of squared distances:
$$
W_n=\frac{1}{n}\sum_{i=1}^n\min_{1\leq j\leq k}\big\Vert x_i-\theta_j\big\Vert_2^2.
$$
We advocate the use of center-based clustering methods as an outcome-blind preliminary filter in knot-point selection. From the perspective of vector quantization, $k$-means converges to an optimal $k$-quantizer under resource constraints, placing knot points where they best represent the spatial support of the covariates. Framing HAL knot screening as a vector quantization problem opens avenues for leveraging signal processing methodologies within the HAL framework to support scalable estimation in higher-dimensional settings.

\subsection{PAM}
Partitioning Around Medoids (PAM) \cite{kaufman_rousseeuw_1990} is a center-based clustering method that partitions observations into groups represented by medoids. Unlike $k$-means, whose cluster centroids need not correspond to observed data points, PAM selects representative observations from the original data.

We consider PAM as an alternative outcome-blind knot selection strategy. As with $k$-means, clustering is applied to the observed predictor values in each subset $s$ using the same cluster budget (Section \ref{sec:budget}), and the resulting medoids serve as knot points in the HAL basis expansion. Because medoids are actual observations, PAM knots form a subset of the observed data ($\Theta \subset \Phi$). PAM minimizes the sum of distances between observations and their nearest medoids rather than the within-cluster sum of squares, which makes it more robust to outliers but generally more computationally expensive than $k$-means.

\subsection{Knot budgets and number of basis functions}\label{sec:budget}
For each subset $s$ of predictors, $k$-means and PAM are applied to the observed values $x_{1,s},\dots,x_{n,s}$ of the predictors in $s$, and the resulting cluster centers ($k$-means) or medoids (PAM) are used as knot points. To compare the screening algorithms at approximately matched design matrix sizes, the numbers of cluster centers for $k$-means and PAM are derived from the basis produced by quantile-based discretization. Let $N_1$, $N_2$, and $N_3$ be the numbers of first-, second-, and third-degree basis functions produced by quantile-based discretization, $d$ the number of predictors, and $d_c$ the number of continuous predictors. The numbers of cluster centers per subset of one, two, and three predictors are
$$
k_1=N_1/d_c,\qquad k_2=N_2\Big/\binom{d}{2},\qquad k_3=N_3\Big/\binom{d}{3},
$$
corresponding to the average number of quantile-based knots per subset of each degree. Non-integer values and subsets with fewer distinct observed points than cluster centers follow the \texttt{hal9001} implementation used in this study. In the simulations, $d_c$ is set to $d$; in the real-world data, $d_c$ is the number of predictors with more than two distinct values in the training data (or $d$ if no continuous predictor is present). Because a uniform budget is applied across subsets of a given degree and binary predictors contribute at most two distinct values, $k$-means and PAM yield slightly fewer basis functions than quantile-based discretization: the same number for $p=1$, 8--10\% fewer for $p=3$, and 2--4\% fewer for $p=5$ in the simulations; and up to 6\% fewer for seven of the nine real-world datasets (11--13\% fewer for bike sharing and 26\% fewer for automobile).

All three knot-screening algorithms are outcome-blind, using only the training covariates; the outcome is used solely when fitting the cross-validated lasso to the resulting basis matrix (and for initial variable screening in two high-dimensional real-world datasets; Section \ref{sec:real}). Throughout, HAL is fitted with zero-order basis functions and a maximum interaction degree of three. Tables report the total number of basis functions (columns of the HAL design matrix passed to \texttt{glmnet}); we focus on predictive accuracy and basis-matrix size rather than wall-clock runtime.

\section{Simulations}\label{sec:sim}
We compare the predictive performance of quantile-based discretization, $k$-means, and PAM across 17 data-generating processes (DGPs) for $X$: the regular design of \cite{hal_2016} (Section \ref{sec:sim-regular}), three correlated designs (Section \ref{sec:sim-corr}), one skewed design with independent variables, and twelve skewed and correlated designs (Section \ref{sec:sim-skew}). Each design is paired with the smooth, jump, and sinusoidal regression functions of \cite{hal_2016} in $p \in \{1, 3, 5\}$ dimensions, where $p=1$ and $p=3$ use the first one or three of the five covariates. In all settings, $Y = f_0(X) + \varepsilon$ with $\varepsilon \sim \mathrm{Normal}(0,1)$ independent of $X$, and $f_0$ is one of the nine regression functions listed in Appendix \ref{app:functions}.

In each of 10 independent replicates per setting, a training set of $n=1000$ observations and an independent test set of 10{,}000 observations (1{,}000 for the independent skewed design) are drawn from the same distribution. All three screening algorithms are applied to the same training split, and predictive accuracy is evaluated by test-set $R^2 = 1 - \sum_i \{Y_i - \hat{f}(X_i)\}^2 / \sum_i (Y_i - \bar{Y})^2$. Under the regular design, each regression function is scaled so that the population optimal $R^2$, $R^2_{\mathrm{opt}} = \Var\{f_0(X)\} / [\Var\{f_0(X)\} + 1]$, is approximately $0.80$ (between $0.799$ and $0.804$ by Monte Carlo). Under the remaining designs, $\Var\{f_0(X)\}$ changes with the covariate distribution ($R^2_{\mathrm{opt}} \in [0.70, 0.82]$ for correlated designs, $[0.80, 0.94]$ for independent skewed, and $[0.45, 0.99]$ for skewed and correlated), so test $R^2$ values are compared across methods within each setting rather than across designs.

For conciseness, we present the regular design (Table \ref{tab:sim-regular}) and the cross-design overview (Table \ref{tab:sim-overview}) in the main text, and defer the full per-design tables for the correlated and skewed settings to Appendix \ref{app:sim-tables}. In all simulation tables, \emph{Avg.\ test $R^2$ (SD)} reports the mean and sample SD over 10 replicates (with the highest mean in bold), and superscripts $^{+}$ ($^{-}$) indicate that the unadjusted 95\% paired $t$-interval for the within-replicate difference against Quantile lies strictly above (below) zero.

\subsection{Regular}\label{sec:sim-regular}
This design follows the original HAL simulation study \cite{hal_2016}, with independent covariates $X_1, X_2\sim\mathrm{Uniform}(-4,4)$, $X_3\sim\mathrm{Bernoulli}(0.5)$, $X_4\sim\mathrm{Normal}(0,1)$, and $X_5\sim\mathrm{Gamma}(\mathrm{shape}=2, \mathrm{rate}=1)$.

\begin{table}[H]
\centering
\caption{Simulation results under the regular design. Bold indicates the highest mean test $R^2$; superscripts $^{+}$ ($^{-}$) denote significantly higher (lower) test $R^2$ than Quantile.}
\label{tab:sim-regular}
\resizebox{\ifdim\width>\textwidth\textwidth\else\width\fi}{!}{%
\begin{tabular}{rlcccccc}
\toprule
 & & \multicolumn{3}{c}{Avg. test $R^2$ (SD)} & \multicolumn{3}{c}{Avg. total number of bases} \\
\cmidrule(lr){3-5}\cmidrule(lr){6-8}
$p$ & Scenario & Quantile & $k$-means & PAM & Quantile & $k$-means & PAM \\
\midrule
1 & smooth & \textbf{0.793} (0.005) & 0.792 (0.005) & 0.793 (0.005) & 50.0 & 50.0 & 50.0 \\
 & jump & 0.761 (0.007) & 0.756 (0.014) & \textbf{0.762} (0.008) & 50.0 & 50.0 & 50.0 \\
 & sin & 0.786 (0.004) & 0.784$^{-}$ (0.004) & \textbf{0.787} (0.004) & 50.0 & 50.0 & 50.0 \\
\midrule
3 & smooth & 0.781 (0.004) & \textbf{0.782} (0.004) & 0.781 (0.004) & 424.8 & 391.9 & 391.9 \\
 & jump & 0.749 (0.008) & \textbf{0.758}$^{+}$ (0.008) & 0.757$^{+}$ (0.006) & 426.7 & 394.2 & 394.2 \\
 & sin & 0.585 (0.018) & \textbf{0.626}$^{+}$ (0.011) & 0.610$^{+}$ (0.019) & 428.0 & 395.0 & 395.0 \\
\midrule
5 & smooth & 0.774 (0.005) & 0.775 (0.005) & \textbf{0.775} (0.005) & 2307.5 & 2259.0 & 2259.0 \\
 & jump & 0.745 (0.009) & 0.763$^{+}$ (0.005) & \textbf{0.764}$^{+}$ (0.007) & 2312.2 & 2263.0 & 2263.0 \\
 & sin & 0.527 (0.019) & \textbf{0.573}$^{+}$ (0.018) & 0.569$^{+}$ (0.024) & 2310.2 & 2262.0 & 2262.0 \\
\bottomrule
\end{tabular}}
\end{table}

As shown in Table \ref{tab:sim-regular}, $k$-means and PAM achieve higher test $R^2$ than quantile-based discretization for the jump and sinusoidal functions when $p > 1$ (gains of $0.009$ to $0.046$) and comparable test $R^2$ for $p=1$ and for the smooth function, while using slightly fewer basis functions.

\subsection{Correlated variables}\label{sec:sim-corr}
To evaluate performance under covariate dependence, we generate $X$ via a Gaussian copula with latent correlation $corr \in \{0.1, 0.5, 0.9\}$:
\begin{align*}
S = (S_1, S_2, S_3, S_4) &\sim \mathrm{Normal}(\mu,\Sigma),\\
X_1 &= 8\Phi(S_1) - 4,\\
X_2 &= 8\Phi(S_2) - 4,\\
X_3 &\sim \mathrm{Bernoulli}(\mathrm{logit}^{-1}(0.25 X_1 - 0.25 X_2)),\\
X_4 &= S_3 + X_1 \times X_3,\\
X_5 &= q_{\gamma}\left(\Phi\left(S_4\right), \mathrm{shape} = 2, \mathrm{rate} = 1\right) + 2,
\end{align*}
where $\mu = (0,0,0,0)$, $\Sigma$ has $1$ on the diagonal and $corr$ off the diagonal, and $\Phi$ is the standard normal CDF. Thus $X_1$ and $X_2$ retain $\mathrm{Uniform}(-4,4)$ margins as in the regular design. Full results and sample correlation matrices are reported in Tables \ref{tab:sim-corr_0.1}--\ref{tab:sim-corr_0.9} and \ref{tab:corr-matrices} (Appendix \ref{app:sim-tables}). As in the regular design, $k$-means and PAM use fewer basis functions yet significantly improve on Quantile in 10 and 8 of the 27 correlated settings, with gains of up to $0.047$ for the jump and sinusoidal functions with $p \in \{3, 5\}$; the few settings in which Quantile is better differ by at most $0.006$.

\subsection{Skewed variables}\label{sec:sim-skew}
We next consider skewed marginal distributions, with training and test sets drawn from the same distribution (no distribution shift). First, we evaluate an independent skewed design (Table \ref{tab:sim-skewed} in Appendix \ref{app:sim-tables}):
\begin{align*}
X_1 &= \mathrm{Gamma}(\mathrm{shape}=1, \mathrm{scale}=2) - 5, \\
X_2 &= \mathrm{Gamma}(\mathrm{shape}=2, \mathrm{scale}=2) - 2, \\
X_3 &\sim \mathrm{Bernoulli}(0.9), \\
X_4 &= \mathrm{Gamma}(\mathrm{shape}=2, \mathrm{scale}=1) \times
\begin{cases}
-1 & \text{if } \mathrm{Uniform}(0,1) > 0.8, \\
1 & \text{otherwise},
\end{cases} \\
X_5 &= \mathrm{Gamma}(\mathrm{shape}=0.5, \mathrm{scale}=1).
\end{align*}

Second, to study the combined effect of skewness and correlation, we replace the uniform margins of the correlated design with standardized Gamma margins of prescribed skewness $skewness \in \{0.5, 1, 2, 3\}$ and latent correlation $corr \in \{0.1, 0.5, 0.9\}$ (Tables \ref{tab:sim-gskew_0.5_corr_0.1}--\ref{tab:sim-gskew_3_corr_0.9} in Appendix \ref{app:sim-tables}). With $S$ as in Section \ref{sec:sim-corr} and $a = 4/skewness^2$, let
\begin{align*}
Z_j = \frac{q_{\gamma}\left(\Phi\left(S_j\right), \mathrm{shape} = a, \mathrm{rate} = 1\right) - a}{\sqrt{a}}, \quad j = 1, 2, 3, 4,
\end{align*}
which has mean $0$, variance $1$, and skewness equal to $skewness$, and set
\begin{align*}
X_1 &= \tfrac{8}{\sqrt{12}}\, Z_1, \qquad
X_2 = \tfrac{8}{\sqrt{12}}\, Z_2, \\
X_3 &\sim \mathrm{Bernoulli}(\mathrm{logit}^{-1}(0.25 X_1 - 0.25 X_2)), \\
X_4 &= Z_3 + X_1 \times X_3, \qquad
X_5 = 4 + \sqrt{2}\, Z_4.
\end{align*}
Because these transformations are monotone, the Spearman correlation among $X_1, X_2, X_5$ remains fixed at $(6/\pi)\arcsin(corr/2)$ across all skewness levels, while $X_1$ and $X_2$ become bounded below by $-(8/\sqrt{12})\allowbreak(2/skewness)$ with increasingly heavy right tails. Consequently, at higher skewness ($skewness \ge 2$), the regression functions are evaluated over a longer right tail and some left-tail jump points (e.g., $x_1 = -3$) are no longer crossed.

\subsection{Summary of simulation results}\label{sec:sim-summary}
Table \ref{tab:sim-overview} summarizes all 17 designs via the paired differences in test $R^2$ relative to quantile-based discretization. Here, \emph{Avg.\ $\Delta R^2$} averages the mean paired difference across the nine settings ($p \in \{1,3,5\}$ and three regression functions), with Welch--Satterthwaite 95\% intervals reflecting Monte Carlo variability, and \emph{Better / worse} counts how many of the nine settings had paired 95\% intervals strictly above or below zero.

\begin{table}[H]
\centering
\caption{Cross-design summary of paired differences in test $R^2$ relative to quantile-based discretization, averaged over the nine settings of each design ($^{*}$: 95\% confidence interval excludes zero).}
\label{tab:sim-overview}
\resizebox{\ifdim\width>\textwidth\textwidth\else\width\fi}{!}{%
\begin{tabular}{lllcccc}
\toprule
 & & & \multicolumn{2}{c}{$k$-means $-$ Quantile} & \multicolumn{2}{c}{PAM $-$ Quantile} \\
\cmidrule(lr){4-5}\cmidrule(lr){6-7}
Design & Skewness & Corr. & Avg. $\Delta R^2$ [95\% CI] & Better / worse & Avg. $\Delta R^2$ [95\% CI] & Better / worse \\
\midrule
Regular & 0 & 0 & $+$0.012$^{*}$ {\footnotesize[$+$0.009, $+$0.015]} & 4 / 1 & $+$0.011$^{*}$ {\footnotesize[$+$0.008, $+$0.013]} & 4 / 0 \\
Correlated & 0 & 0.1 & $+$0.010$^{*}$ {\footnotesize[$+$0.007, $+$0.013]} & 4 / 3 & $+$0.009$^{*}$ {\footnotesize[$+$0.006, $+$0.011]} & 3 / 1 \\
Correlated & 0 & 0.5 & $+$0.009$^{*}$ {\footnotesize[$+$0.007, $+$0.012]} & 3 / 0 & $+$0.007$^{*}$ {\footnotesize[$+$0.005, $+$0.010]} & 3 / 1 \\
Correlated & 0 & 0.9 & $+$0.003$^{*}$ {\footnotesize[$+$0.001, $+$0.005]} & 3 / 0 & $+$0.006$^{*}$ {\footnotesize[$+$0.004, $+$0.008]} & 2 / 0 \\
\midrule
Skewed (indep.) & 2 & 0 & $+$0.029$^{*}$ {\footnotesize[$+$0.023, $+$0.035]} & 6 / 0 & $+$0.035$^{*}$ {\footnotesize[$+$0.029, $+$0.042]} & 8 / 0 \\
\midrule
Skewed + corr. & 0.5 & 0.1 & $+$0.033$^{*}$ {\footnotesize[$+$0.030, $+$0.036]} & 7 / 1 & $+$0.034$^{*}$ {\footnotesize[$+$0.030, $+$0.037]} & 8 / 0 \\
Skewed + corr. & 0.5 & 0.5 & $+$0.030$^{*}$ {\footnotesize[$+$0.027, $+$0.034]} & 7 / 0 & $+$0.035$^{*}$ {\footnotesize[$+$0.031, $+$0.038]} & 8 / 0 \\
Skewed + corr. & 0.5 & 0.9 & $+$0.024$^{*}$ {\footnotesize[$+$0.021, $+$0.028]} & 6 / 0 & $+$0.028$^{*}$ {\footnotesize[$+$0.025, $+$0.032]} & 7 / 0 \\
Skewed + corr. & 1 & 0.1 & $+$0.047$^{*}$ {\footnotesize[$+$0.037, $+$0.057]} & 6 / 0 & $+$0.053$^{*}$ {\footnotesize[$+$0.042, $+$0.064]} & 7 / 0 \\
Skewed + corr. & 1 & 0.5 & $+$0.045$^{*}$ {\footnotesize[$+$0.040, $+$0.050]} & 7 / 0 & $+$0.051$^{*}$ {\footnotesize[$+$0.046, $+$0.056]} & 8 / 0 \\
Skewed + corr. & 1 & 0.9 & $+$0.038$^{*}$ {\footnotesize[$+$0.033, $+$0.042]} & 6 / 0 & $+$0.043$^{*}$ {\footnotesize[$+$0.039, $+$0.047]} & 7 / 0 \\
Skewed + corr. & 2 & 0.1 & $+$0.061$^{*}$ {\footnotesize[$+$0.053, $+$0.069]} & 7 / 0 & $+$0.073$^{*}$ {\footnotesize[$+$0.065, $+$0.081]} & 8 / 0 \\
Skewed + corr. & 2 & 0.5 & $+$0.054$^{*}$ {\footnotesize[$+$0.046, $+$0.063]} & 5 / 1 & $+$0.074$^{*}$ {\footnotesize[$+$0.068, $+$0.081]} & 7 / 0 \\
Skewed + corr. & 2 & 0.9 & $+$0.056$^{*}$ {\footnotesize[$+$0.048, $+$0.063]} & 5 / 0 & $+$0.075$^{*}$ {\footnotesize[$+$0.069, $+$0.081]} & 6 / 0 \\
Skewed + corr. & 3 & 0.1 & $+$0.110$^{*}$ {\footnotesize[$+$0.073, $+$0.147]} & 7 / 1 & $+$0.132$^{*}$ {\footnotesize[$+$0.099, $+$0.165]} & 8 / 0 \\
Skewed + corr. & 3 & 0.5 & $+$0.084$^{*}$ {\footnotesize[$+$0.074, $+$0.095]} & 7 / 0 & $+$0.110$^{*}$ {\footnotesize[$+$0.101, $+$0.119]} & 9 / 0 \\
Skewed + corr. & 3 & 0.9 & $+$0.068$^{*}$ {\footnotesize[$+$0.060, $+$0.077]} & 7 / 0 & $+$0.097$^{*}$ {\footnotesize[$+$0.088, $+$0.106]} & 8 / 0 \\
\bottomrule
\end{tabular}}
\end{table}

Two clear patterns emerge across the 153 simulation settings:

\begin{enumerate}
\item \textbf{Uniform margins (regular and correlated designs):} Even with uniform margins, clustering-based screening outperforms Quantile in 14 ($k$-means) and 12 (PAM) of the 36 settings, almost all for the jump and sinusoidal functions with $p \ge 3$ (gains up to $0.047$; design averages $0.003$ to $0.012$), while Quantile is better in only 4 and 2 settings, by at most $0.006$.
\item \textbf{Skewed margins (independent and correlated):} Clustering-based screening shows substantial and consistent gains. In the independent skewed design, $k$-means and PAM outperform Quantile in 6 and 8 of the 9 settings (average gains $0.029$ and $0.035$). Across the 108 skewed-and-correlated settings, PAM outperforms Quantile in 91 settings and is never significantly worse, while $k$-means outperforms Quantile in 77 settings and is worse in only 3 (all with $p=1$). PAM's average gain is at least as large as that of $k$-means in all 13 skewed designs.
\end{enumerate}

The increase in average gain with skewness is driven primarily by the smooth regression function: averaged over $p$ and $corr$, PAM's gain on the smooth function rises from $0.053$ at $skewness=0.5$ to $0.110$, $0.188$, and $0.275$ at $skewness=1, 2, 3$ ($k$-means: $0.045, 0.096, 0.152, 0.232$), whereas gains on the jump and sinusoidal functions remain relatively stable ($0.017$--$0.031$ for PAM). Mechanistically, this behavior aligns with the vector-quantization view in Section \ref{sec:screening}: quantile discretization allocates knots evenly in probability mass, leaving the sparse but wide right tail of a skewed distribution severely under-covered. Because the smooth quadratic function varies rapidly in the tail, coarse tail spacing produces large piecewise-constant approximation errors—in extreme cases causing Quantile to yield negative test $R^2$ (e.g., $-0.83$ and $-0.16$ in two replicates at $skewness=3, corr=0.1, p=3$, smooth; mean $0.34$, median $0.54$, versus medians $0.78$ for $k$-means and $0.84$ for PAM). By contrast, $k$-means and PAM minimize spatial distortion and naturally allocate knots across the extended tail.

In contrast to skewness, stronger covariate correlation does not amplify the advantage of clustering and generally reduces it for $p \in \{3, 5\}$ (for $p=1$, $X_1$'s marginal distribution does not depend on $corr$). For instance, at $skewness=3$, the average gain of $k$-means declines from $0.134$ ($corr=0.1$) to $0.097$ ($corr=0.5$) and $0.075$ ($corr=0.9$), with a similar trend for PAM ($0.146, 0.109, 0.098$). Notably, despite producing slightly smaller design matrices, $k$-means and PAM select more active basis functions at the cross-validated penalty (for $p=5$, averaging 164 and 169 nonzero coefficients respectively, versus 136 for Quantile), i.e., the lasso retains a larger fraction of their basis functions.

\section{Real World Data}\label{sec:real}
\subsection{Datasets and evaluation protocol}\label{sec:real-setup}
We evaluate the three knot-screening algorithms on nine publicly available benchmark datasets summarized in Table \ref{table:data}, with sample sizes ranging from $159$ to $731$ and $5$ to $122$ predictors. Four of these datasets (\texttt{cpu}, \texttt{laheart}, \texttt{oecdpanel}, and \texttt{pima}) were also analyzed in the original HAL paper \cite{hal_2016}. Predictive performance is measured by mean squared error (MSE).

\begin{table}[ht]
\centering
\caption{Real-world datasets analyzed with HAL. $n$: complete-case sample size; $d$: number of predictors after preprocessing, with the number of continuous predictors (more than two distinct values) in parentheses.}
\label{table:data}
\begin{tabular}{lrrl}
\toprule
Name & $n$ & $d$ & Outcome modeled \\
\midrule
automobile \cite{misc_automobile_10} & 159 & 55 (15) & symboling (insurance risk rating) \\
communities and crime \cite{misc_communities_and_crime_183} & 319 & 122 (122) & violent crimes per 100{,}000 population (normalized) \\
liver disorders \cite{misc_liver_disorders_60} & 345 & 5 (5) & drinks (half-pint equivalents per day) \\
bike sharing \cite{misc_bike_sharing_275} & 731 & 11 (8) & daily count of rental bikes \\
cpu \cite{computer_hardware_29} & 209 & 6 (6) & published relative performance \\
laheart \cite{afifi_azen_1979} & 200 & 11 (11) & death (0/1) \\
real estate \cite{real_estate_valuation_477} & 414 & 6 (6) & house price per unit area \\
oecdpanel \cite{liu_stengos_1999,hayfield_racine_2008} & 616 & 6 (5) & humancap (school enrolment) \\
pima \cite{smith_pima_1988} & 392 & 7 (7) & diabetes (0/1) \\
\bottomrule
\end{tabular}
\end{table}

All datasets are analyzed as complete cases after removing constant predictors; categorical predictors stored as text are dummy-coded, and integer-coded predictors (e.g., \texttt{season} and \texttt{weekday} in \texttt{bike sharing}) are treated as numeric. Binary outcomes (\texttt{laheart} and \texttt{pima}) are modeled under squared-error loss. Performance is estimated via 10-fold cross-validation repeated 10 times with matched folds across all methods and knot budgets. All screening and HAL fitting occur strictly within training folds. For the two higher-dimensional datasets (\texttt{automobile} and \texttt{communities and crime}), we first apply variable-level screening within each training fold using cross-validated lasso, retaining predictors with nonzero coefficients at the minimum-CV penalty (or the first predictor(s) entering the path if none are selected).

To compare a clustering method against quantile discretization at a given knot budget while accounting for overlapping training folds, we use the Nadeau--Bengio corrected repeated cross-validation $t$-test \cite{nadeau_bengio_2003,bouckaert_frank_2004} over the $J=100$ fold-level MSE differences:
$$
\bar\delta \pm t_{0.975,\,J-1}\sqrt{\Big(\frac{1}{J}+\frac{n_{\mathrm{test}}}{n_{\mathrm{train}}}\Big)s^2},
$$
where $\bar\delta$ and $s^2$ are the mean and sample variance of the fold-level differences and $n_{\mathrm{test}}/n_{\mathrm{train}} \approx 1/9$. To assess systematic differences across datasets, we summarize each dataset by its mean percentage MSE difference from Quantile averaged over the nine knot budgets and apply exact two-sided Wilcoxon signed-rank and sign tests \cite{demsar_2006}, alongside the Spearman correlation with average predictor skewness.

Dataset-specific descriptions are summarized below, and full per-dataset result tables (Tables \ref{tab:real-cpu}--\ref{tab:real-real_estate}) are provided in Appendix \ref{app:real-tables}.

\begin{itemize}
\item \textbf{cpu} \cite{computer_hardware_29}: Relative performance (\texttt{prp}, range 6--1{,}150, strongly right-skewed) of 209 processors predicted from six continuous hardware attributes (\texttt{myct}, \texttt{mmin}, \texttt{mmax}, \texttt{cach}, \texttt{chmin}, \texttt{chmax}), exhibiting the highest predictor skewness among all datasets (average skewness 3.06).
\item \textbf{automobile} \cite{misc_automobile_10}: Insurance risk rating (\texttt{symboling}, $-2$ to $3$) for 159 cars with complete records, predicted from 15 numeric variables and 9 categorical variables coded into 40 indicators (55 predictors total, screened fold-wise by lasso).
\item \textbf{communities and crime} \cite{misc_communities_and_crime_183}: Normalized violent crimes per 100{,}000 population across 319 US communities with complete Census and LEMAS police records, predicted from 122 normalized numeric variables (screened fold-wise by lasso).
\item \textbf{pima} \cite{smith_pima_1988}: Diabetes status (0/1; 130 cases) among 392 women of Pima Indian heritage, predicted from seven clinical measurements (parity, glucose, blood pressure, skinfold thickness, insulin, BMI, age).
\item \textbf{oecdpanel} \cite{liu_stengos_1999,hayfield_racine_2008}: Cross-country growth panel of 88 countries across seven five-year periods (616 observations), modeling log secondary school enrolment (\texttt{humancap}) from six macroeconomic predictors.
\item \textbf{bike sharing} \cite{misc_bike_sharing_275}: Daily rental bike counts (\texttt{cnt}, range 22--8{,}714) over 731 days in Washington, D.C., predicted from 11 calendar and weather variables.
\item \textbf{liver disorders} \cite{misc_liver_disorders_60}: Daily alcohol consumption (\texttt{drinks}, range 0--20) for 345 male individuals, predicted from five blood-test markers (\texttt{mcv}, \texttt{alkphos}, \texttt{sgpt}, \texttt{sgot}, \texttt{gammagt}).
\item \textbf{laheart} \cite{afifi_azen_1979}: Mortality indicator (0/1; 64 deaths) for 200 individuals, predicted from age, blood pressure, height, weight, cholesterol, socioeconomic status, and four additional measurements (\texttt{sbp2}, \texttt{dbp2}, \texttt{chol2}, \texttt{wt2}).
\item \textbf{real estate} \cite{real_estate_valuation_477}: House price per unit area (range 7.6--117.5) for 414 transactions in New Taipei City, Taiwan, predicted from six temporal, structural, and geographic variables.
\end{itemize}

\subsection{Summary across data sets}\label{sec:real-summary}
Clustering-based screening delivers its largest gain on \texttt{cpu}, the dataset with the most heavily skewed predictors (average skewness 3.06; Table \ref{tab:disc}). There, $k$-means and PAM reduce the average MSE by 35\% and 36\% and achieve a lower MSE than Quantile in all 10 CV repeats at every knot budget. They are also insensitive to the knot budget (average MSE between 2{,}396 and 2{,}664 at all nine budgets), whereas the MSE of Quantile rises to 6{,}431 at the smallest budget $(20, 10, 5)$ (3{,}700--3{,}827 at the other budgets). With the smallest budget, $k$-means and PAM are therefore more accurate than Quantile with the largest budget, while using about 7\% fewer basis functions (1{,}474 versus 1{,}582).

Across the nine datasets (Table \ref{tab:real-across}), PAM achieves a lower average MSE than Quantile on 7 datasets and $k$-means on 5, and PAM reduces the MSE by 8.3\% on \texttt{automobile}; on the remaining datasets, both clustering methods stay within 2.5\% of Quantile. Because the gains on \texttt{cpu} are driven by folds in which Quantile makes very large errors, the fold-level differences are highly variable, and neither the per-budget comparisons nor the across-dataset tests reach statistical significance.

\begin{table}[H]
\centering
\caption{Cross-dataset summary of relative differences in cross-validated MSE compared to quantile-based discretization across the nine real-world datasets.}
\label{tab:real-across}
\begin{tabular}{lcccccc}
\toprule
 & Data sets with & Median & Wilcoxon & Sign test & \multicolumn{2}{c}{Spearman (skewness, diff.)} \\
\cmidrule(lr){6-7}
Method & lower MSE & difference & $p$ & $p$ & $\rho$ & $p$ \\
\midrule
$k$-means & 5 of 9 & $-$0.4\% & 0.50 & 1.00 & $-$0.40 & 0.29 \\
PAM & 7 of 9 & $-$0.9\% & 0.13 & 0.18 & $-$0.08 & 0.84 \\
\bottomrule
\end{tabular}
\end{table}

\section{Discussion}\label{sec:discussion}
In this paper, we investigated outcome-blind knot point screening for HAL, comparing \texttt{hal9001}'s default quantile-based discretization against two center-based clustering quantizers ($k$-means and PAM) under matched basis-budget rules. With slightly smaller HAL design matrices, clustering-based screening significantly improved test $R^2$ in 97 ($k$-means) and 111 (PAM) of the 153 simulation settings, with design-average gains of up to $0.13$ under strongly skewed covariates, and was worse in only 7 and 2 settings, by at most $0.017$. On real data, it reduced the cross-validated MSE by 35--36\% on the most skewed dataset and remained competitive on the others.

\begin{table}[ht]
\centering
\caption{Real-world datasets: frequency of a lower average MSE for $k$-means or PAM relative to Quantile across the nine knot budgets, alongside skewness and correlation characteristics of the continuous predictors.}
\label{tab:disc}
\resizebox{\textwidth}{!}{%
\begin{tabular}{lrrrr}
\toprule
\textbf{Dataset} & \textbf{$k$-means or PAM better} & \textbf{Avg.\ skewness} & \textbf{$|$Skewness$|$ $>$ 0.5} & \textbf{$|$Correlation$|$ $>$ 0.5} \\
\midrule
cpu & 100\% & 3.06 & 100\% & 53\% \\
automobile & 100\% & 0.71 & 73\% & 44\% \\
communities and crime & 89\% & 0.89 & 77\% & 11\% \\
pima & 100\% & 0.88 & 71\% & 14\% \\
oecdpanel & 100\% & $-0.40$ & 40\% & 20\% \\
bike sharing & 100\% & 0.17 & 25\% & 11\% \\
liver disorders & 0\% & 1.70 & 80\% & 30\% \\
laheart & 78\% & 0.44 & 45\% & 7\% \\
real estate & 100\% & 0.10 & 33\% & 20\% \\
\bottomrule
\end{tabular}}
\end{table}

Table \ref{tab:disc} summarizes the structural properties of the continuous predictors in the nine real-world datasets alongside the proportion of knot budgets in which at least one clustering method achieved a lower average MSE than Quantile (because this column counts the budgets at which the better of the two methods beats Quantile, it should be read together with the per-method results in Table \ref{tab:real-across}). The other columns use the sample skewness $m_3/s^3$ of each continuous predictor, where $m_3$ is the third central sample moment and $s$ the sample standard deviation: \emph{Avg.\ skewness} is its mean over the continuous predictors and \emph{$|$Skewness$|$ $>$ 0.5} the percentage of continuous predictors with absolute skewness above 0.5; \emph{$|$Correlation$|$ $>$ 0.5} is the percentage of pairs of continuous predictors with absolute Pearson correlation above 0.5. All are computed on the full preprocessed dataset (for \texttt{automobile} and \texttt{communities and crime}, before the fold-wise lasso screening). In our simulations, the advantage of $k$-means and PAM over quantile discretization is largest under skewed covariates paired with smooth regression functions, where equal-probability quantiles under-represent the sparse right tail. In the real-world benchmarks, the most skewed dataset (\texttt{cpu}) shows by far the largest MSE reduction (35--36\%), in line with this mechanism. With only nine datasets, the rank correlation between average skewness and the change in MSE (Spearman $\rho = -0.40$ for $k$-means and $-0.08$ for PAM) is not significant, and larger benchmark collections are needed to quantify how the gain depends on skewness. In the simulations, stronger covariate correlation did not amplify the advantage of clustering-based screening and, for $p \in \{3, 5\}$, generally reduced it (Section \ref{sec:sim-summary}).

From a practical standpoint, our results support treating clustering-based knot screening as a strong, drop-in alternative to quantile discretization within HAL's modular screening framework---ideally selected via cross-validation rather than chosen solely on the basis of marginal skewness---and, as the \texttt{cpu} results show, it can be markedly more robust than quantile discretization when the knot budget is small. Between the two clustering methods, $k$-means and PAM deliver comparable predictive accuracy across both simulated and real-world benchmarks; PAM gives the largest gains under skewed covariates, while $k$-means is computationally cheaper (Section \ref{sec:screening}) and thus a natural default for larger datasets.

This study focuses on zero-order HAL basis functions and predictive accuracy at the knot-point screening level. Several natural extensions remain for future work: benchmarking end-to-end wall-clock runtime (including the clustering step) on larger sample sizes and higher-order spline HAL bases, integrating knot-point screening with automated variable- and basis-level screening, and evaluating how clustering-screened HAL performs as a nuisance parameter estimator within semiparametric causal inference procedures such as TMLE.

The code to reproduce all simulations and real-world data analyses is available at \url{https://github.com/yadiweng/hal-knot-screening}. The $k$-means and PAM screening options were run with a development version of \texttt{hal9001} (version 0.4.5; GitHub repository \texttt{tq21/hal9001}, branch \texttt{kmeans-knot-screen}, commit \texttt{a81af69}).

\printbibliography

\appendix
\section{Regression functions}\label{app:functions}
The nine regression functions $f_0$ follow \cite{hal_2016} as implemented in our simulation code. In all 17 designs, $Y=f_0(X)+\varepsilon$ with $\varepsilon\sim\mathrm{Normal}(0,1)$ independent of $X$, and $f_0$ depends on $x_1$, $(x_1,x_2,x_3)$, and $(x_1,\dots,x_5)$ for $p=1, 3, 5$, respectively.

For $p=1$:
\begin{align*}
f_{\mathrm{smooth}}(x) &= 0.05x_1-0.42x_1^2,\\
f_{\mathrm{jump}}(x) &= -2.7I(x_1<-3)+2.5I(x_1>-2)-2I(x_1>0)+4I(x_1>2)-3I(x_1>3),\\
f_{\mathrm{sin}}(x) &= 2\sin(\pi|x_1|/2)+2\cos(\pi|x_1|/2).
\end{align*}
For $p=3$:
\begin{align*}
f_{\mathrm{smooth}}(x) &= x_1/15-0.28x_1^2+0.5x_2+0.25x_2x_3,\\
f_{\mathrm{jump}}(x) &= -2I(x_1<-3)x_3+2.5I(x_1>-2)-2I(x_1>0)+2.5I(x_1>2)x_3-2.5I(x_1>3)\\
&\quad +I(x_2>-1)-4I(x_2>1)x_3+2I(x_2>3),\\
f_{\mathrm{sin}}(x) &= x_3\{4\sin(\pi|x_1|/2)I(x_2<0)+4.1\cos(\pi|x_1|/2)I(x_2>0)\}.
\end{align*}
For $p=5$:
\begin{align*}
f_{\mathrm{smooth}}(x) &= x_1/10-0.3x_1^2+0.25x_2+0.5x_2x_3-0.5x_4+0.04x_5^2-0.1x_5,\\
f_{\mathrm{jump}}(x) &= -I(x_1<-3)x_3+0.5I(x_1>-2)-I(x_1>0)+2I(x_1>2)x_3-3I(x_1>3)+1.5I(x_2>-1)\\
&\quad -5I(x_2>1)x_3+2I(x_2>3)+2I(x_4<0)-I(x_5>5)-I(x_4<0)I(x_1<0)+2x_3,\\
f_{\mathrm{sin}}(x) &= x_3\{3.75\sin(\pi|x_1|/2)I(x_2<0)+4\cos(\pi|x_1|/2)I(x_2>0)\}+\sin(\pi x_4)x_5/10+\cos(|x_4-x_5|)x_3.
\end{align*}
Under the regular design, the population optimal $R^2$ of each setting is approximately $0.80$ (between $0.799$ and $0.804$ by Monte Carlo).

\section{Detailed simulation results}\label{app:sim-tables}
This appendix contains the full per-design simulation tables for the correlated designs (Tables \ref{tab:sim-corr_0.1}--\ref{tab:sim-corr_0.9}), their sample covariate correlation matrices (Table \ref{tab:corr-matrices}), the independent skewed design (Table \ref{tab:sim-skewed}), and the twelve skewed-and-correlated designs (Tables \ref{tab:sim-gskew_0.5_corr_0.1}--\ref{tab:sim-gskew_3_corr_0.9}). The columns are as in Table \ref{tab:sim-regular}: \emph{Avg.\ test $R^2$ (SD)} is the mean and SD over the 10 replicates, with the highest mean in bold, and a superscript $^{+}$ ($^{-}$) marks a significantly higher (lower) test $R^2$ than Quantile (paired 95\% $t$-interval); \emph{Avg.\ total number of bases} is the number of columns of the HAL design matrix, averaged over the 10 replicates.

\begin{table}[H]
\centering
\caption{Simulation results under the correlated design ($corr = 0.1$).}
\label{tab:sim-corr_0.1}
\resizebox{\ifdim\width>\textwidth\textwidth\else\width\fi}{!}{%
\begin{tabular}{rlcccccc}
\toprule
 & & \multicolumn{3}{c}{Avg. test $R^2$ (SD)} & \multicolumn{3}{c}{Avg. total number of bases} \\
\cmidrule(lr){3-5}\cmidrule(lr){6-8}
$p$ & Scenario & Quantile & $k$-means & PAM & Quantile & $k$-means & PAM \\
\midrule
1 & smooth & \textbf{0.795} (0.006) & 0.795 (0.006) & 0.794 (0.007) & 50.0 & 50.0 & 50.0 \\
 & jump & 0.764 (0.012) & 0.758$^{-}$ (0.012) & \textbf{0.768} (0.012) & 50.0 & 50.0 & 50.0 \\
 & sin & \textbf{0.787} (0.003) & 0.786$^{-}$ (0.004) & 0.787 (0.005) & 50.0 & 50.0 & 50.0 \\
\midrule
3 & smooth & \textbf{0.787} (0.004) & 0.786$^{-}$ (0.004) & 0.786$^{-}$ (0.004) & 424.5 & 391.5 & 391.5 \\
 & jump & 0.724 (0.011) & \textbf{0.738}$^{+}$ (0.010) & 0.736 (0.007) & 425.5 & 392.7 & 392.7 \\
 & sin & 0.602 (0.015) & \textbf{0.623}$^{+}$ (0.017) & 0.608 (0.018) & 426.7 & 393.7 & 393.7 \\
\midrule
5 & smooth & 0.790 (0.004) & 0.791 (0.004) & \textbf{0.792}$^{+}$ (0.003) & 2222.2 & 2173.0 & 2173.0 \\
 & jump & 0.730 (0.018) & \textbf{0.750}$^{+}$ (0.007) & 0.747$^{+}$ (0.008) & 2225.6 & 2176.0 & 2176.0 \\
 & sin & 0.507 (0.013) & \textbf{0.554}$^{+}$ (0.016) & 0.546$^{+}$ (0.016) & 2228.8 & 2177.0 & 2177.0 \\
\bottomrule
\end{tabular}}
\end{table}

\begin{table}[H]
\centering
\caption{Simulation results under the correlated design ($corr = 0.5$).}
\label{tab:sim-corr_0.5}
\resizebox{\ifdim\width>\textwidth\textwidth\else\width\fi}{!}{%
\begin{tabular}{rlcccccc}
\toprule
 & & \multicolumn{3}{c}{Avg. test $R^2$ (SD)} & \multicolumn{3}{c}{Avg. total number of bases} \\
\cmidrule(lr){3-5}\cmidrule(lr){6-8}
$p$ & Scenario & Quantile & $k$-means & PAM & Quantile & $k$-means & PAM \\
\midrule
1 & smooth & \textbf{0.793} (0.004) & 0.793 (0.004) & 0.793 (0.004) & 50.0 & 50.0 & 50.0 \\
 & jump & 0.761 (0.010) & 0.759 (0.012) & \textbf{0.764} (0.009) & 50.0 & 50.0 & 50.0 \\
 & sin & 0.786 (0.004) & 0.785 (0.003) & \textbf{0.786} (0.003) & 50.0 & 50.0 & 50.0 \\
\midrule
3 & smooth & \textbf{0.792} (0.004) & 0.791 (0.003) & 0.791$^{-}$ (0.004) & 404.4 & 371.1 & 371.1 \\
 & jump & 0.717 (0.011) & 0.726 (0.006) & \textbf{0.727}$^{+}$ (0.011) & 405.1 & 371.9 & 371.9 \\
 & sin & 0.617 (0.016) & \textbf{0.638}$^{+}$ (0.021) & 0.625 (0.017) & 406.7 & 373.4 & 373.4 \\
\midrule
5 & smooth & 0.739 (0.006) & \textbf{0.740} (0.006) & 0.740 (0.008) & 2055.0 & 2005.0 & 2005.0 \\
 & jump & 0.739 (0.014) & \textbf{0.756}$^{+}$ (0.007) & 0.754$^{+}$ (0.006) & 2047.7 & 1998.0 & 1998.0 \\
 & sin & 0.525 (0.017) & \textbf{0.564}$^{+}$ (0.020) & 0.554$^{+}$ (0.010) & 2059.6 & 2014.0 & 2014.0 \\
\bottomrule
\end{tabular}}
\end{table}

\begin{table}[H]
\centering
\caption{Simulation results under the correlated design ($corr = 0.9$).}
\label{tab:sim-corr_0.9}
\resizebox{\ifdim\width>\textwidth\textwidth\else\width\fi}{!}{%
\begin{tabular}{rlcccccc}
\toprule
 & & \multicolumn{3}{c}{Avg. test $R^2$ (SD)} & \multicolumn{3}{c}{Avg. total number of bases} \\
\cmidrule(lr){3-5}\cmidrule(lr){6-8}
$p$ & Scenario & Quantile & $k$-means & PAM & Quantile & $k$-means & PAM \\
\midrule
1 & smooth & 0.791 (0.003) & 0.791 (0.003) & \textbf{0.792} (0.003) & 50.0 & 50.0 & 50.0 \\
 & jump & 0.762 (0.007) & 0.757 (0.014) & \textbf{0.765} (0.015) & 50.0 & 50.0 & 50.0 \\
 & sin & 0.787 (0.004) & 0.786 (0.005) & \textbf{0.787} (0.004) & 50.0 & 50.0 & 50.0 \\
\midrule
3 & smooth & 0.799 (0.004) & \textbf{0.800} (0.003) & 0.799 (0.005) & 323.4 & 290.0 & 290.0 \\
 & jump & 0.700 (0.013) & 0.696 (0.012) & \textbf{0.708} (0.015) & 323.0 & 290.5 & 290.5 \\
 & sin & 0.680 (0.005) & 0.692$^{+}$ (0.010) & \textbf{0.697}$^{+}$ (0.012) & 321.7 & 288.5 & 288.5 \\
\midrule
5 & smooth & 0.654 (0.006) & 0.655 (0.004) & \textbf{0.656} (0.008) & 1457.9 & 1409.0 & 1409.0 \\
 & jump & 0.752 (0.009) & \textbf{0.765}$^{+}$ (0.011) & 0.761 (0.009) & 1468.6 & 1422.0 & 1422.0 \\
 & sin & 0.648 (0.011) & 0.660$^{+}$ (0.009) & \textbf{0.661}$^{+}$ (0.006) & 1449.4 & 1401.0 & 1401.0 \\
\bottomrule
\end{tabular}}
\end{table}

\begin{table}[H]
\centering
\caption{Sample correlation matrices of $X_1$ to $X_5$ in the correlated designs (one simulated dataset for each value of $corr$).}
\label{tab:corr-matrices}
\small
$corr=0.1$
\[
\begin{array}{|c|c|c|c|c|c|}
\hline
    & X_1 & X_2 & X_3 & X_4 & X_5 \\
\hline
X_1 & 1.000 & 0.048 & 0.236 & 0.644 & 0.020 \\
X_2 & 0.048 & 1.000 & -0.256 & 0.082 & 0.076 \\
X_3 & 0.236 & -0.256 & 1.000 & 0.137 & -0.038 \\
X_4 & 0.644 & 0.082 & 0.137 & 1.000 & 0.059 \\
X_5 & 0.020 & 0.076 & -0.038 & 0.059 & 1.000 \\
\hline
\end{array}
\]
$corr=0.5$
\[
\begin{array}{|c|c|c|c|c|c|}
\hline
    & X_1 & X_2 & X_3 & X_4 & X_5 \\
\hline
X_1 & 1.000 & 0.481 & 0.113 & 0.738 & 0.244 \\
X_2 & 0.481 & 1.000 & -0.162 & 0.452 & 0.268 \\
X_3 & 0.113 & -0.162 & 1.000 & 0.079 & 0.034 \\
X_4 & 0.738 & 0.452 & 0.079 & 1.000 & 0.259 \\
X_5 & 0.244 & 0.268 & 0.034 & 0.259 & 1.000 \\
\hline
\end{array}
\]
$corr=0.9$
\[
\begin{array}{|c|c|c|c|c|c|}
\hline
    & X_1 & X_2 & X_3 & X_4 & X_5 \\
\hline
X_1 & 1.000 & 0.885 & 0.035 & 0.865 & 0.507 \\
X_2 & 0.885 & 1.000 & -0.030 & 0.800 & 0.505 \\
X_3 & 0.035 & -0.030 & 1.000 & -0.007 & -0.000 \\
X_4 & 0.865 & 0.800 & -0.007 & 1.000 & 0.463 \\
X_5 & 0.507 & 0.505 & -0.000 & 0.463 & 1.000 \\
\hline
\end{array}
\]
\end{table}

\begin{table}[H]
\centering
\caption{Simulation results under the independent skewed design.}
\label{tab:sim-skewed}
\resizebox{\ifdim\width>\textwidth\textwidth\else\width\fi}{!}{%
\begin{tabular}{rlcccccc}
\toprule
 & & \multicolumn{3}{c}{Avg. test $R^2$ (SD)} & \multicolumn{3}{c}{Avg. total number of bases} \\
\cmidrule(lr){3-5}\cmidrule(lr){6-8}
$p$ & Scenario & Quantile & $k$-means & PAM & Quantile & $k$-means & PAM \\
\midrule
1 & smooth & 0.801 (0.058) & 0.884$^{+}$ (0.023) & \textbf{0.903}$^{+}$ (0.023) & 50.0 & 50.0 & 50.0 \\
 & jump & 0.778 (0.012) & 0.776 (0.013) & \textbf{0.787}$^{+}$ (0.011) & 50.0 & 50.0 & 50.0 \\
 & sin & 0.762 (0.007) & 0.765 (0.007) & \textbf{0.776}$^{+}$ (0.007) & 50.0 & 50.0 & 50.0 \\
\midrule
3 & smooth & 0.838 (0.029) & 0.874$^{+}$ (0.015) & \textbf{0.884}$^{+}$ (0.013) & 424.0 & 391.2 & 391.2 \\
 & jump & 0.810 (0.010) & 0.823$^{+}$ (0.005) & \textbf{0.824}$^{+}$ (0.008) & 422.2 & 389.5 & 389.5 \\
 & sin & 0.740 (0.021) & \textbf{0.744} (0.012) & 0.738 (0.016) & 421.8 & 389.0 & 389.0 \\
\midrule
5 & smooth & 0.789 (0.053) & 0.874$^{+}$ (0.022) & \textbf{0.884}$^{+}$ (0.016) & 2282.2 & 2233.0 & 2233.0 \\
 & jump & 0.747 (0.027) & 0.772$^{+}$ (0.021) & \textbf{0.774}$^{+}$ (0.018) & 2287.8 & 2237.0 & 2237.0 \\
 & sin & 0.712 (0.020) & \textbf{0.729}$^{+}$ (0.022) & 0.725$^{+}$ (0.016) & 2284.4 & 2237.0 & 2237.0 \\
\bottomrule
\end{tabular}}
\end{table}

\begin{table}[H]
\centering
\caption{Simulation results under the skewed and correlated design ($\text{skewness} = 0.5$, $corr = 0.1$).}
\label{tab:sim-gskew_0.5_corr_0.1}
\resizebox{\ifdim\width>\textwidth\textwidth\else\width\fi}{!}{%
\begin{tabular}{rlcccccc}
\toprule
 & & \multicolumn{3}{c}{Avg. test $R^2$ (SD)} & \multicolumn{3}{c}{Avg. total number of bases} \\
\cmidrule(lr){3-5}\cmidrule(lr){6-8}
$p$ & Scenario & Quantile & $k$-means & PAM & Quantile & $k$-means & PAM \\
\midrule
1 & smooth & 0.829 (0.021) & 0.880$^{+}$ (0.018) & \textbf{0.894}$^{+}$ (0.013) & 50.0 & 50.0 & 50.0 \\
 & jump & \textbf{0.709} (0.014) & 0.700$^{-}$ (0.011) & 0.703 (0.009) & 50.0 & 50.0 & 50.0 \\
 & sin & 0.774 (0.007) & 0.773 (0.006) & \textbf{0.782}$^{+}$ (0.003) & 50.0 & 50.0 & 50.0 \\
\midrule
3 & smooth & 0.786 (0.022) & 0.830$^{+}$ (0.008) & \textbf{0.836}$^{+}$ (0.012) & 423.8 & 390.6 & 390.6 \\
 & jump & 0.701 (0.015) & \textbf{0.718}$^{+}$ (0.010) & 0.717$^{+}$ (0.009) & 424.3 & 391.8 & 391.8 \\
 & sin & 0.516 (0.009) & \textbf{0.578}$^{+}$ (0.011) & 0.557$^{+}$ (0.013) & 423.8 & 390.6 & 390.6 \\
\midrule
5 & smooth & 0.812 (0.022) & 0.851$^{+}$ (0.020) & \textbf{0.851}$^{+}$ (0.020) & 2208.2 & 2159.0 & 2159.0 \\
 & jump & 0.707 (0.021) & \textbf{0.733}$^{+}$ (0.011) & 0.732$^{+}$ (0.013) & 2220.1 & 2171.0 & 2171.0 \\
 & sin & 0.424 (0.016) & \textbf{0.493}$^{+}$ (0.021) & 0.489$^{+}$ (0.022) & 2225.0 & 2178.0 & 2178.0 \\
\bottomrule
\end{tabular}}
\end{table}

\begin{table}[H]
\centering
\caption{Simulation results under the skewed and correlated design ($\text{skewness} = 0.5$, $corr = 0.5$).}
\label{tab:sim-gskew_0.5_corr_0.5}
\resizebox{\ifdim\width>\textwidth\textwidth\else\width\fi}{!}{%
\begin{tabular}{rlcccccc}
\toprule
 & & \multicolumn{3}{c}{Avg. test $R^2$ (SD)} & \multicolumn{3}{c}{Avg. total number of bases} \\
\cmidrule(lr){3-5}\cmidrule(lr){6-8}
$p$ & Scenario & Quantile & $k$-means & PAM & Quantile & $k$-means & PAM \\
\midrule
1 & smooth & 0.836 (0.014) & 0.885$^{+}$ (0.008) & \textbf{0.900}$^{+}$ (0.003) & 50.0 & 50.0 & 50.0 \\
 & jump & 0.701 (0.010) & 0.696 (0.019) & \textbf{0.703} (0.012) & 50.0 & 50.0 & 50.0 \\
 & sin & 0.777 (0.002) & 0.778 (0.005) & \textbf{0.787}$^{+}$ (0.004) & 50.0 & 50.0 & 50.0 \\
\midrule
3 & smooth & 0.784 (0.017) & 0.817$^{+}$ (0.012) & \textbf{0.827}$^{+}$ (0.004) & 406.3 & 373.2 & 373.2 \\
 & jump & 0.713 (0.012) & \textbf{0.724}$^{+}$ (0.013) & 0.719$^{+}$ (0.012) & 406.7 & 374.1 & 374.1 \\
 & sin & 0.542 (0.019) & 0.591$^{+}$ (0.015) & \textbf{0.598}$^{+}$ (0.018) & 406.0 & 373.2 & 373.2 \\
\midrule
5 & smooth & 0.768 (0.022) & 0.827$^{+}$ (0.013) & \textbf{0.832}$^{+}$ (0.011) & 2056.7 & 2006.0 & 2006.0 \\
 & jump & 0.742 (0.014) & \textbf{0.763}$^{+}$ (0.007) & 0.758$^{+}$ (0.008) & 2057.8 & 2007.0 & 2007.0 \\
 & sin & 0.457 (0.025) & 0.509$^{+}$ (0.014) & \textbf{0.510}$^{+}$ (0.016) & 2072.5 & 2023.0 & 2023.0 \\
\bottomrule
\end{tabular}}
\end{table}

\begin{table}[H]
\centering
\caption{Simulation results under the skewed and correlated design ($\text{skewness} = 0.5$, $corr = 0.9$).}
\label{tab:sim-gskew_0.5_corr_0.9}
\resizebox{\ifdim\width>\textwidth\textwidth\else\width\fi}{!}{%
\begin{tabular}{rlcccccc}
\toprule
 & & \multicolumn{3}{c}{Avg. test $R^2$ (SD)} & \multicolumn{3}{c}{Avg. total number of bases} \\
\cmidrule(lr){3-5}\cmidrule(lr){6-8}
$p$ & Scenario & Quantile & $k$-means & PAM & Quantile & $k$-means & PAM \\
\midrule
1 & smooth & 0.834 (0.014) & 0.879$^{+}$ (0.014) & \textbf{0.895}$^{+}$ (0.005) & 50.0 & 50.0 & 50.0 \\
 & jump & \textbf{0.704} (0.006) & 0.702 (0.016) & 0.699 (0.009) & 50.0 & 50.0 & 50.0 \\
 & sin & 0.773 (0.004) & 0.774 (0.008) & \textbf{0.782}$^{+}$ (0.007) & 50.0 & 50.0 & 50.0 \\
\midrule
3 & smooth & 0.765 (0.030) & 0.807$^{+}$ (0.013) & \textbf{0.811}$^{+}$ (0.006) & 320.4 & 287.6 & 287.6 \\
 & jump & 0.702 (0.017) & \textbf{0.708} (0.012) & 0.703 (0.013) & 321.1 & 287.9 & 287.9 \\
 & sin & 0.635 (0.013) & 0.667$^{+}$ (0.015) & \textbf{0.680}$^{+}$ (0.013) & 322.2 & 289.3 & 289.3 \\
\midrule
5 & smooth & 0.747 (0.009) & 0.792$^{+}$ (0.015) & \textbf{0.796}$^{+}$ (0.015) & 1456.8 & 1409.0 & 1409.0 \\
 & jump & 0.768 (0.014) & \textbf{0.781}$^{+}$ (0.009) & 0.781$^{+}$ (0.010) & 1462.5 & 1412.0 & 1412.0 \\
 & sin & 0.622 (0.009) & 0.659$^{+}$ (0.007) & \textbf{0.659}$^{+}$ (0.004) & 1439.0 & 1392.0 & 1392.0 \\
\bottomrule
\end{tabular}}
\end{table}

\begin{table}[H]
\centering
\caption{Simulation results under the skewed and correlated design ($\text{skewness} = 1$, $corr = 0.1$).}
\label{tab:sim-gskew_1_corr_0.1}
\resizebox{\ifdim\width>\textwidth\textwidth\else\width\fi}{!}{%
\begin{tabular}{rlcccccc}
\toprule
 & & \multicolumn{3}{c}{Avg. test $R^2$ (SD)} & \multicolumn{3}{c}{Avg. total number of bases} \\
\cmidrule(lr){3-5}\cmidrule(lr){6-8}
$p$ & Scenario & Quantile & $k$-means & PAM & Quantile & $k$-means & PAM \\
\midrule
1 & smooth & 0.798 (0.033) & 0.876$^{+}$ (0.022) & \textbf{0.903}$^{+}$ (0.017) & 50.0 & 50.0 & 50.0 \\
 & jump & 0.672 (0.011) & 0.661 (0.013) & \textbf{0.673} (0.014) & 50.0 & 50.0 & 50.0 \\
 & sin & 0.777 (0.006) & 0.779 (0.011) & \textbf{0.789}$^{+}$ (0.006) & 50.0 & 50.0 & 50.0 \\
\midrule
3 & smooth & 0.745 (0.067) & 0.839$^{+}$ (0.030) & \textbf{0.856}$^{+}$ (0.012) & 423.7 & 390.7 & 390.7 \\
 & jump & 0.706 (0.007) & \textbf{0.710} (0.009) & 0.709 (0.009) & 422.8 & 389.6 & 389.6 \\
 & sin & 0.518 (0.014) & \textbf{0.564}$^{+}$ (0.016) & 0.559$^{+}$ (0.012) & 424.2 & 391.4 & 391.4 \\
\midrule
5 & smooth & 0.743 (0.117) & 0.873$^{+}$ (0.024) & \textbf{0.877}$^{+}$ (0.020) & 2211.9 & 2165.0 & 2165.0 \\
 & jump & 0.713 (0.011) & \textbf{0.732}$^{+}$ (0.010) & 0.731$^{+}$ (0.008) & 2209.1 & 2161.0 & 2161.0 \\
 & sin & 0.425 (0.022) & \textbf{0.486}$^{+}$ (0.026) & 0.477$^{+}$ (0.028) & 2219.6 & 2171.0 & 2171.0 \\
\bottomrule
\end{tabular}}
\end{table}

\begin{table}[H]
\centering
\caption{Simulation results under the skewed and correlated design ($\text{skewness} = 1$, $corr = 0.5$).}
\label{tab:sim-gskew_1_corr_0.5}
\resizebox{\ifdim\width>\textwidth\textwidth\else\width\fi}{!}{%
\begin{tabular}{rlcccccc}
\toprule
 & & \multicolumn{3}{c}{Avg. test $R^2$ (SD)} & \multicolumn{3}{c}{Avg. total number of bases} \\
\cmidrule(lr){3-5}\cmidrule(lr){6-8}
$p$ & Scenario & Quantile & $k$-means & PAM & Quantile & $k$-means & PAM \\
\midrule
1 & smooth & 0.765 (0.026) & 0.870$^{+}$ (0.023) & \textbf{0.904}$^{+}$ (0.016) & 50.0 & 50.0 & 50.0 \\
 & jump & 0.670 (0.013) & 0.664 (0.011) & \textbf{0.671} (0.009) & 50.0 & 50.0 & 50.0 \\
 & sin & 0.774 (0.006) & 0.777 (0.011) & \textbf{0.790}$^{+}$ (0.004) & 50.0 & 50.0 & 50.0 \\
\midrule
3 & smooth & 0.717 (0.028) & 0.807$^{+}$ (0.026) & \textbf{0.819}$^{+}$ (0.030) & 406.2 & 373.2 & 373.2 \\
 & jump & 0.701 (0.012) & \textbf{0.713}$^{+}$ (0.008) & 0.712$^{+}$ (0.010) & 406.1 & 373.1 & 373.1 \\
 & sin & 0.557 (0.013) & \textbf{0.586}$^{+}$ (0.022) & 0.585$^{+}$ (0.020) & 405.8 & 372.3 & 372.3 \\
\midrule
5 & smooth & 0.747 (0.040) & \textbf{0.852}$^{+}$ (0.021) & 0.851$^{+}$ (0.030) & 2052.5 & 2003.0 & 2003.0 \\
 & jump & 0.735 (0.018) & 0.751$^{+}$ (0.010) & \textbf{0.753}$^{+}$ (0.013) & 2055.8 & 2005.0 & 2005.0 \\
 & sin & 0.457 (0.016) & \textbf{0.506}$^{+}$ (0.026) & 0.497$^{+}$ (0.023) & 2045.8 & 1995.0 & 1995.0 \\
\bottomrule
\end{tabular}}
\end{table}

\begin{table}[H]
\centering
\caption{Simulation results under the skewed and correlated design ($\text{skewness} = 1$, $corr = 0.9$).}
\label{tab:sim-gskew_1_corr_0.9}
\resizebox{\ifdim\width>\textwidth\textwidth\else\width\fi}{!}{%
\begin{tabular}{rlcccccc}
\toprule
 & & \multicolumn{3}{c}{Avg. test $R^2$ (SD)} & \multicolumn{3}{c}{Avg. total number of bases} \\
\cmidrule(lr){3-5}\cmidrule(lr){6-8}
$p$ & Scenario & Quantile & $k$-means & PAM & Quantile & $k$-means & PAM \\
\midrule
1 & smooth & 0.778 (0.029) & 0.870$^{+}$ (0.022) & \textbf{0.885}$^{+}$ (0.022) & 50.0 & 50.0 & 50.0 \\
 & jump & \textbf{0.665} (0.018) & 0.655 (0.026) & 0.664 (0.014) & 50.0 & 50.0 & 50.0 \\
 & sin & 0.776 (0.005) & 0.779 (0.006) & \textbf{0.790}$^{+}$ (0.005) & 50.0 & 50.0 & 50.0 \\
\midrule
3 & smooth & 0.712 (0.019) & 0.791$^{+}$ (0.017) & \textbf{0.806}$^{+}$ (0.018) & 322.5 & 289.8 & 289.8 \\
 & jump & 0.688 (0.008) & \textbf{0.693} (0.013) & 0.689 (0.011) & 322.5 & 289.8 & 289.8 \\
 & sin & 0.649 (0.008) & 0.680$^{+}$ (0.019) & \textbf{0.687}$^{+}$ (0.012) & 321.6 & 288.6 & 288.6 \\
\midrule
5 & smooth & 0.710 (0.032) & 0.802$^{+}$ (0.013) & \textbf{0.802}$^{+}$ (0.022) & 1445.1 & 1396.0 & 1396.0 \\
 & jump & 0.764 (0.006) & \textbf{0.776}$^{+}$ (0.006) & 0.773$^{+}$ (0.007) & 1443.3 & 1392.0 & 1392.0 \\
 & sin & 0.610 (0.012) & \textbf{0.647}$^{+}$ (0.011) & 0.643$^{+}$ (0.012) & 1453.3 & 1405.0 & 1405.0 \\
\bottomrule
\end{tabular}}
\end{table}

\begin{table}[H]
\centering
\caption{Simulation results under the skewed and correlated design ($\text{skewness} = 2$, $corr = 0.1$).}
\label{tab:sim-gskew_2_corr_0.1}
\resizebox{\ifdim\width>\textwidth\textwidth\else\width\fi}{!}{%
\begin{tabular}{rlcccccc}
\toprule
 & & \multicolumn{3}{c}{Avg. test $R^2$ (SD)} & \multicolumn{3}{c}{Avg. total number of bases} \\
\cmidrule(lr){3-5}\cmidrule(lr){6-8}
$p$ & Scenario & Quantile & $k$-means & PAM & Quantile & $k$-means & PAM \\
\midrule
1 & smooth & 0.711 (0.046) & 0.869$^{+}$ (0.052) & \textbf{0.897}$^{+}$ (0.051) & 50.0 & 50.0 & 50.0 \\
 & jump & 0.523 (0.015) & 0.515 (0.018) & \textbf{0.525} (0.020) & 50.0 & 50.0 & 50.0 \\
 & sin & 0.755 (0.006) & 0.748 (0.011) & \textbf{0.773}$^{+}$ (0.006) & 50.0 & 50.0 & 50.0 \\
\midrule
3 & smooth & 0.638 (0.085) & 0.793$^{+}$ (0.092) & \textbf{0.825}$^{+}$ (0.097) & 422.6 & 389.5 & 389.5 \\
 & jump & 0.686 (0.013) & \textbf{0.698}$^{+}$ (0.009) & 0.695$^{+}$ (0.010) & 424.1 & 390.9 & 390.9 \\
 & sin & 0.530 (0.012) & 0.568$^{+}$ (0.012) & \textbf{0.574}$^{+}$ (0.015) & 422.8 & 389.9 & 389.9 \\
\midrule
5 & smooth & 0.712 (0.044) & \textbf{0.870}$^{+}$ (0.048) & 0.868$^{+}$ (0.041) & 2199.6 & 2150.0 & 2150.0 \\
 & jump & 0.696 (0.011) & 0.711$^{+}$ (0.011) & \textbf{0.713}$^{+}$ (0.010) & 2202.7 & 2154.0 & 2154.0 \\
 & sin & 0.474 (0.021) & 0.503$^{+}$ (0.020) & \textbf{0.510}$^{+}$ (0.020) & 2203.8 & 2154.0 & 2154.0 \\
\bottomrule
\end{tabular}}
\end{table}

\begin{table}[H]
\centering
\caption{Simulation results under the skewed and correlated design ($\text{skewness} = 2$, $corr = 0.5$).}
\label{tab:sim-gskew_2_corr_0.5}
\resizebox{\ifdim\width>\textwidth\textwidth\else\width\fi}{!}{%
\begin{tabular}{rlcccccc}
\toprule
 & & \multicolumn{3}{c}{Avg. test $R^2$ (SD)} & \multicolumn{3}{c}{Avg. total number of bases} \\
\cmidrule(lr){3-5}\cmidrule(lr){6-8}
$p$ & Scenario & Quantile & $k$-means & PAM & Quantile & $k$-means & PAM \\
\midrule
1 & smooth & 0.705 (0.067) & 0.854$^{+}$ (0.064) & \textbf{0.903}$^{+}$ (0.051) & 50.0 & 50.0 & 50.0 \\
 & jump & \textbf{0.532} (0.016) & 0.515$^{-}$ (0.025) & 0.524 (0.015) & 50.0 & 50.0 & 50.0 \\
 & sin & 0.758 (0.007) & 0.759 (0.008) & \textbf{0.775}$^{+}$ (0.006) & 50.0 & 50.0 & 50.0 \\
\midrule
3 & smooth & 0.634 (0.063) & 0.807$^{+}$ (0.043) & \textbf{0.841}$^{+}$ (0.047) & 407.1 & 373.8 & 373.8 \\
 & jump & 0.677 (0.010) & \textbf{0.682} (0.011) & 0.677 (0.009) & 405.4 & 372.4 & 372.4 \\
 & sin & 0.529 (0.021) & 0.541$^{+}$ (0.022) & \textbf{0.555}$^{+}$ (0.020) & 407.8 & 374.8 & 374.8 \\
\midrule
5 & smooth & 0.703 (0.050) & 0.820$^{+}$ (0.107) & \textbf{0.872}$^{+}$ (0.040) & 2053.6 & 2004.0 & 2004.0 \\
 & jump & 0.724 (0.007) & 0.732 (0.011) & \textbf{0.739}$^{+}$ (0.006) & 2045.1 & 1997.0 & 1997.0 \\
 & sin & 0.458 (0.014) & 0.498$^{+}$ (0.020) & \textbf{0.501}$^{+}$ (0.024) & 2042.6 & 1994.0 & 1994.0 \\
\bottomrule
\end{tabular}}
\end{table}

\begin{table}[H]
\centering
\caption{Simulation results under the skewed and correlated design ($\text{skewness} = 2$, $corr = 0.9$).}
\label{tab:sim-gskew_2_corr_0.9}
\resizebox{\ifdim\width>\textwidth\textwidth\else\width\fi}{!}{%
\begin{tabular}{rlcccccc}
\toprule
 & & \multicolumn{3}{c}{Avg. test $R^2$ (SD)} & \multicolumn{3}{c}{Avg. total number of bases} \\
\cmidrule(lr){3-5}\cmidrule(lr){6-8}
$p$ & Scenario & Quantile & $k$-means & PAM & Quantile & $k$-means & PAM \\
\midrule
1 & smooth & 0.726 (0.035) & 0.877$^{+}$ (0.030) & \textbf{0.917}$^{+}$ (0.029) & 50.0 & 50.0 & 50.0 \\
 & jump & \textbf{0.529} (0.013) & 0.515 (0.018) & 0.522 (0.014) & 50.0 & 50.0 & 50.0 \\
 & sin & 0.759 (0.008) & 0.760 (0.010) & \textbf{0.772}$^{+}$ (0.005) & 50.0 & 50.0 & 50.0 \\
\midrule
3 & smooth & 0.615 (0.062) & 0.768$^{+}$ (0.062) & \textbf{0.815}$^{+}$ (0.055) & 320.4 & 287.1 & 287.1 \\
 & jump & \textbf{0.663} (0.012) & 0.662 (0.012) & 0.657 (0.017) & 322.6 & 289.6 & 289.6 \\
 & sin & 0.575 (0.021) & 0.604$^{+}$ (0.012) & \textbf{0.618}$^{+}$ (0.009) & 324.0 & 290.8 & 290.8 \\
\midrule
5 & smooth & 0.657 (0.065) & 0.813$^{+}$ (0.032) & \textbf{0.851}$^{+}$ (0.040) & 1459.3 & 1410.0 & 1410.0 \\
 & jump & 0.766 (0.009) & 0.766 (0.010) & \textbf{0.773} (0.008) & 1444.5 & 1395.0 & 1395.0 \\
 & sin & 0.572 (0.014) & 0.599$^{+}$ (0.019) & \textbf{0.612}$^{+}$ (0.010) & 1446.1 & 1399.0 & 1399.0 \\
\bottomrule
\end{tabular}}
\end{table}

\begin{table}[H]
\centering
\caption{Simulation results under the skewed and correlated design ($\text{skewness} = 3$, $corr = 0.1$).}
\label{tab:sim-gskew_3_corr_0.1}
\resizebox{\ifdim\width>\textwidth\textwidth\else\width\fi}{!}{%
\begin{tabular}{rlcccccc}
\toprule
 & & \multicolumn{3}{c}{Avg. test $R^2$ (SD)} & \multicolumn{3}{c}{Avg. total number of bases} \\
\cmidrule(lr){3-5}\cmidrule(lr){6-8}
$p$ & Scenario & Quantile & $k$-means & PAM & Quantile & $k$-means & PAM \\
\midrule
1 & smooth & 0.629 (0.038) & 0.844$^{+}$ (0.050) & \textbf{0.887}$^{+}$ (0.056) & 50.0 & 50.0 & 50.0 \\
 & jump & 0.387 (0.030) & 0.367 (0.043) & \textbf{0.409} (0.014) & 50.0 & 50.0 & 50.0 \\
 & sin & 0.613 (0.013) & 0.601$^{-}$ (0.019) & \textbf{0.641}$^{+}$ (0.009) & 50.0 & 50.0 & 50.0 \\
\midrule
3 & smooth & 0.340 (0.475) & 0.797$^{+}$ (0.074) & \textbf{0.828}$^{+}$ (0.099) & 422.1 & 388.9 & 388.9 \\
 & jump & 0.626 (0.009) & 0.644$^{+}$ (0.015) & \textbf{0.662}$^{+}$ (0.007) & 421.0 & 388.2 & 388.2 \\
 & sin & 0.555 (0.022) & 0.578$^{+}$ (0.025) & \textbf{0.582}$^{+}$ (0.027) & 423.7 & 390.2 & 390.2 \\
\midrule
5 & smooth & 0.630 (0.085) & 0.867$^{+}$ (0.040) & \textbf{0.878}$^{+}$ (0.048) & 2185.3 & 2136.0 & 2136.0 \\
 & jump & 0.655 (0.012) & 0.693$^{+}$ (0.012) & \textbf{0.702}$^{+}$ (0.007) & 2186.1 & 2135.0 & 2135.0 \\
 & sin & 0.525 (0.015) & 0.554$^{+}$ (0.020) & \textbf{0.557}$^{+}$ (0.019) & 2183.1 & 2133.0 & 2133.0 \\
\bottomrule
\end{tabular}}
\end{table}

\begin{table}[H]
\centering
\caption{Simulation results under the skewed and correlated design ($\text{skewness} = 3$, $corr = 0.5$).}
\label{tab:sim-gskew_3_corr_0.5}
\resizebox{\ifdim\width>\textwidth\textwidth\else\width\fi}{!}{%
\begin{tabular}{rlcccccc}
\toprule
 & & \multicolumn{3}{c}{Avg. test $R^2$ (SD)} & \multicolumn{3}{c}{Avg. total number of bases} \\
\cmidrule(lr){3-5}\cmidrule(lr){6-8}
$p$ & Scenario & Quantile & $k$-means & PAM & Quantile & $k$-means & PAM \\
\midrule
1 & smooth & 0.641 (0.080) & 0.821$^{+}$ (0.097) & \textbf{0.918}$^{+}$ (0.053) & 50.0 & 50.0 & 50.0 \\
 & jump & 0.392 (0.025) & 0.395 (0.039) & \textbf{0.414}$^{+}$ (0.016) & 50.0 & 50.0 & 50.0 \\
 & sin & 0.609 (0.007) & 0.606 (0.016) & \textbf{0.641}$^{+}$ (0.006) & 50.0 & 50.0 & 50.0 \\
\midrule
3 & smooth & 0.519 (0.130) & 0.772$^{+}$ (0.133) & \textbf{0.809}$^{+}$ (0.132) & 406.6 & 373.6 & 373.6 \\
 & jump & 0.601 (0.010) & 0.621$^{+}$ (0.013) & \textbf{0.639}$^{+}$ (0.013) & 406.5 & 373.4 & 373.4 \\
 & sin & 0.552 (0.017) & 0.574$^{+}$ (0.013) & \textbf{0.580}$^{+}$ (0.016) & 405.9 & 372.7 & 372.7 \\
\midrule
5 & smooth & 0.647 (0.046) & 0.865$^{+}$ (0.063) & \textbf{0.876}$^{+}$ (0.061) & 2031.2 & 1984.0 & 1984.0 \\
 & jump & 0.683 (0.011) & 0.718$^{+}$ (0.011) & \textbf{0.722}$^{+}$ (0.010) & 2028.0 & 1980.0 & 1980.0 \\
 & sin & 0.532 (0.018) & 0.563$^{+}$ (0.018) & \textbf{0.565}$^{+}$ (0.017) & 2042.7 & 1996.0 & 1996.0 \\
\bottomrule
\end{tabular}}
\end{table}

\begin{table}[H]
\centering
\caption{Simulation results under the skewed and correlated design ($\text{skewness} = 3$, $corr = 0.9$).}
\label{tab:sim-gskew_3_corr_0.9}
\resizebox{\ifdim\width>\textwidth\textwidth\else\width\fi}{!}{%
\begin{tabular}{rlcccccc}
\toprule
 & & \multicolumn{3}{c}{Avg. test $R^2$ (SD)} & \multicolumn{3}{c}{Avg. total number of bases} \\
\cmidrule(lr){3-5}\cmidrule(lr){6-8}
$p$ & Scenario & Quantile & $k$-means & PAM & Quantile & $k$-means & PAM \\
\midrule
1 & smooth & 0.661 (0.066) & 0.842$^{+}$ (0.047) & \textbf{0.893}$^{+}$ (0.049) & 50.0 & 50.0 & 50.0 \\
 & jump & 0.392 (0.028) & 0.375 (0.039) & \textbf{0.410} (0.018) & 50.0 & 50.0 & 50.0 \\
 & sin & 0.613 (0.020) & 0.613 (0.024) & \textbf{0.647}$^{+}$ (0.008) & 50.0 & 50.0 & 50.0 \\
\midrule
3 & smooth & 0.535 (0.056) & 0.709$^{+}$ (0.074) & \textbf{0.788}$^{+}$ (0.095) & 322.1 & 289.1 & 289.1 \\
 & jump & 0.568 (0.011) & 0.588$^{+}$ (0.017) & \textbf{0.604}$^{+}$ (0.015) & 320.1 & 287.2 & 287.2 \\
 & sin & 0.607 (0.014) & 0.633$^{+}$ (0.006) & \textbf{0.643}$^{+}$ (0.012) & 321.4 & 288.6 & 288.6 \\
\midrule
5 & smooth & 0.633 (0.071) & 0.810$^{+}$ (0.094) & \textbf{0.837}$^{+}$ (0.093) & 1451.5 & 1403.0 & 1403.0 \\
 & jump & 0.728 (0.003) & 0.756$^{+}$ (0.007) & \textbf{0.757}$^{+}$ (0.006) & 1444.5 & 1395.0 & 1395.0 \\
 & sin & 0.629 (0.015) & 0.656$^{+}$ (0.015) & \textbf{0.659}$^{+}$ (0.015) & 1445.9 & 1396.0 & 1396.0 \\
\bottomrule
\end{tabular}}
\end{table}

\section{Detailed real-world data results}\label{app:real-tables}
This appendix contains the full repeated 10-fold cross-validation tables for each of the nine real-world datasets (Tables \ref{tab:real-cpu}--\ref{tab:real-real_estate}). \emph{Number of knots} gives the first-, second- and third-degree knot budgets of the quantile-based discretization (Section \ref{sec:screening}); \emph{Avg.\ MSE (SD)} is the mean and SD over the 10 repeats of the cross-validated MSE, with the lowest mean in bold; and \emph{Avg.\ total number of bases} is the number of columns of the HAL design matrix, averaged over the 100 fits (10 folds $\times$ 10 repeats). No difference from Quantile was significant by the corrected test of Section \ref{sec:real-setup}, so no $^{+}$/$^{-}$ markers appear.

\begin{table}[H]
\centering
\caption{Repeated 10-fold cross-validation results for the \texttt{cpu} dataset.}
\label{tab:real-cpu}
\resizebox{\ifdim\width>\textwidth\textwidth\else\width\fi}{!}{%
\begin{tabular}{lcccccc}
\toprule
 & \multicolumn{3}{c}{Avg. MSE (SD)} & \multicolumn{3}{c}{Avg. total number of bases} \\
\cmidrule(lr){2-4}\cmidrule(lr){5-7}
Number of knots & Quantile & $k$-means & PAM & Quantile & $k$-means & PAM \\
\midrule
20, 10, 5 & 6431 (1147) & \textbf{2496} (604) & 2561 (562) & 1494.6 & 1474.4 & 1474.4 \\
30, 10, 5 & 3700 (863) & 2664 (842) & \textbf{2494} (606) & 1546.6 & 1518.2 & 1518.2 \\
40, 10, 5 & 3827 (943) & 2576 (580) & \textbf{2499} (687) & 1556.6 & 1523.9 & 1523.9 \\
50, 10, 5 & 3770 (995) & 2582 (808) & \textbf{2396} (574) & 1561.0 & 1525.9 & 1525.9 \\
60, 10, 5 & 3758 (1004) & \textbf{2493} (677) & 2560 (577) & 1582.3 & 1532.8 & 1532.9 \\
70, 10, 5 & 3759 (1059) & 2613 (937) & \textbf{2565} (537) & 1582.3 & 1532.9 & 1532.9 \\
80, 10, 5 & 3723 (986) & \textbf{2518} (615) & 2542 (586) & 1582.3 & 1532.9 & 1532.9 \\
90, 10, 5 & 3763 (966) & 2563 (558) & \textbf{2527} (558) & 1582.3 & 1532.8 & 1532.9 \\
100, 10, 5 & 3739 (1031) & \textbf{2400} (583) & 2435 (574) & 1582.3 & 1532.9 & 1532.9 \\
\bottomrule
\end{tabular}}
\end{table}

\begin{table}[H]
\centering
\caption{Repeated 10-fold cross-validation results for the \texttt{automobile} dataset.}
\label{tab:real-automobile}
\resizebox{\ifdim\width>\textwidth\textwidth\else\width\fi}{!}{%
\begin{tabular}{lcccccc}
\toprule
 & \multicolumn{3}{c}{Avg. MSE (SD)} & \multicolumn{3}{c}{Avg. total number of bases} \\
\cmidrule(lr){2-4}\cmidrule(lr){5-7}
Number of knots & Quantile & $k$-means & PAM & Quantile & $k$-means & PAM \\
\midrule
20, 10, 5 & 0.3444 (0.0236) & 0.3362 (0.0322) & \textbf{0.3140} (0.0297) & 15295.6 & 11269.5 & 11269.5 \\
30, 10, 5 & 0.3533 (0.0174) & 0.3460 (0.0445) & \textbf{0.3220} (0.0264) & 15325.0 & 11297.4 & 11297.4 \\
40, 10, 5 & 0.3481 (0.0152) & 0.3401 (0.0251) & \textbf{0.3162} (0.0326) & 15373.6 & 11327.6 & 11327.6 \\
50, 10, 5 & 0.3460 (0.0245) & 0.3275 (0.0384) & \textbf{0.3142} (0.0310) & 15391.4 & 11332.9 & 11332.9 \\
60, 10, 5 & 0.3502 (0.0222) & 0.3334 (0.0388) & \textbf{0.3185} (0.0253) & 15398.6 & 11335.2 & 11335.2 \\
70, 10, 5 & 0.3418 (0.0144) & 0.3434 (0.0294) & \textbf{0.3135} (0.0282) & 15402.0 & 11336.1 & 11336.1 \\
80, 10, 5 & 0.3474 (0.0248) & 0.3340 (0.0397) & \textbf{0.3280} (0.0286) & 15404.8 & 11336.8 & 11336.8 \\
90, 10, 5 & 0.3415 (0.0197) & 0.3433 (0.0263) & \textbf{0.3134} (0.0354) & 15407.9 & 11337.5 & 11337.5 \\
100, 10, 5 & 0.3395 (0.0215) & 0.3433 (0.0311) & \textbf{0.3151} (0.0263) & 15410.5 & 11338.0 & 11338.0 \\
\bottomrule
\end{tabular}}
\end{table}

\begin{table}[H]
\centering
\caption{Repeated 10-fold cross-validation results for the \texttt{communities and crime} dataset.}
\label{tab:real-communities_crime}
\resizebox{\ifdim\width>\textwidth\textwidth\else\width\fi}{!}{%
\begin{tabular}{lcccccc}
\toprule
 & \multicolumn{3}{c}{Avg. MSE (SD)} & \multicolumn{3}{c}{Avg. total number of bases} \\
\cmidrule(lr){2-4}\cmidrule(lr){5-7}
Number of knots & Quantile & $k$-means & PAM & Quantile & $k$-means & PAM \\
\midrule
20, 10, 5 & 0.0296 (0.0011) & \textbf{0.0287} (0.0008) & 0.0288 (0.0010) & 50002.3 & 49633.1 & 49633.1 \\
30, 10, 5 & 0.0291 (0.0010) & 0.0290 (0.0009) & \textbf{0.0289} (0.0008) & 50126.4 & 49750.5 & 49750.5 \\
40, 10, 5 & 0.0289 (0.0011) & 0.0293 (0.0009) & \textbf{0.0289} (0.0008) & 50228.0 & 49847.1 & 49847.1 \\
50, 10, 5 & 0.0290 (0.0010) & 0.0289 (0.0011) & \textbf{0.0288} (0.0007) & 50344.1 & 49956.6 & 49956.6 \\
60, 10, 5 & 0.0289 (0.0011) & 0.0291 (0.0010) & \textbf{0.0286} (0.0008) & 50420.6 & 50029.3 & 50029.3 \\
70, 10, 5 & 0.0288 (0.0012) & 0.0292 (0.0010) & \textbf{0.0287} (0.0009) & 50571.7 & 50163.2 & 50163.2 \\
80, 10, 5 & 0.0288 (0.0011) & 0.0290 (0.0007) & \textbf{0.0288} (0.0007) & 50688.4 & 50260.4 & 50260.4 \\
90, 10, 5 & 0.0287 (0.0011) & 0.0289 (0.0009) & \textbf{0.0287} (0.0009) & 50771.8 & 50321.5 & 50321.5 \\
100, 10, 5 & \textbf{0.0286} (0.0011) & 0.0290 (0.0007) & 0.0288 (0.0009) & 50772.4 & 50321.8 & 50321.8 \\
\bottomrule
\end{tabular}}
\end{table}

\begin{table}[H]
\centering
\caption{Repeated 10-fold cross-validation results for the \texttt{pima} dataset.}
\label{tab:real-pima}
\resizebox{\ifdim\width>\textwidth\textwidth\else\width\fi}{!}{%
\begin{tabular}{lcccccc}
\toprule
 & \multicolumn{3}{c}{Avg. MSE (SD)} & \multicolumn{3}{c}{Avg. total number of bases} \\
\cmidrule(lr){2-4}\cmidrule(lr){5-7}
Number of knots & Quantile & $k$-means & PAM & Quantile & $k$-means & PAM \\
\midrule
20, 10, 5 & 0.1551 (0.0020) & \textbf{0.1497} (0.0023) & 0.1524 (0.0020) & 4535.8 & 4503.1 & 4503.1 \\
30, 10, 5 & 0.1551 (0.0028) & \textbf{0.1502} (0.0032) & 0.1519 (0.0026) & 4582.3 & 4544.9 & 4544.9 \\
40, 10, 5 & 0.1522 (0.0034) & \textbf{0.1506} (0.0021) & 0.1512 (0.0016) & 4635.1 & 4589.5 & 4589.5 \\
50, 10, 5 & 0.1525 (0.0028) & \textbf{0.1490} (0.0040) & 0.1512 (0.0017) & 4695.6 & 4636.6 & 4636.6 \\
60, 10, 5 & 0.1525 (0.0036) & \textbf{0.1488} (0.0017) & 0.1517 (0.0015) & 4722.6 & 4653.6 & 4653.6 \\
70, 10, 5 & 0.1528 (0.0031) & \textbf{0.1488} (0.0023) & 0.1520 (0.0027) & 4747.5 & 4666.8 & 4666.8 \\
80, 10, 5 & 0.1527 (0.0033) & \textbf{0.1506} (0.0046) & 0.1511 (0.0019) & 4769.4 & 4676.2 & 4676.2 \\
90, 10, 5 & 0.1522 (0.0032) & \textbf{0.1500} (0.0022) & 0.1520 (0.0023) & 4790.8 & 4685.4 & 4685.4 \\
100, 10, 5 & 0.1526 (0.0032) & \textbf{0.1494} (0.0031) & 0.1512 (0.0019) & 4808.4 & 4692.9 & 4692.9 \\
\bottomrule
\end{tabular}}
\end{table}

\begin{table}[H]
\centering
\caption{Repeated 10-fold cross-validation results for the \texttt{oecdpanel} dataset.}
\label{tab:real-oecdpanel}
\resizebox{\ifdim\width>\textwidth\textwidth\else\width\fi}{!}{%
\begin{tabular}{lcccccc}
\toprule
 & \multicolumn{3}{c}{Avg. MSE (SD)} & \multicolumn{3}{c}{Avg. total number of bases} \\
\cmidrule(lr){2-4}\cmidrule(lr){5-7}
Number of knots & Quantile & $k$-means & PAM & Quantile & $k$-means & PAM \\
\midrule
20, 10, 5 & 0.2163 (0.0041) & 0.2114 (0.0033) & \textbf{0.2110} (0.0061) & 2173.4 & 2100.3 & 2100.3 \\
30, 10, 5 & 0.2147 (0.0059) & 0.2155 (0.0037) & \textbf{0.2099} (0.0059) & 2213.4 & 2132.3 & 2132.3 \\
40, 10, 5 & 0.2141 (0.0063) & 0.2123 (0.0058) & \textbf{0.2099} (0.0057) & 2253.4 & 2164.3 & 2164.3 \\
50, 10, 5 & 0.2133 (0.0063) & 0.2128 (0.0035) & \textbf{0.2100} (0.0057) & 2293.4 & 2196.3 & 2196.3 \\
60, 10, 5 & 0.2127 (0.0052) & 0.2138 (0.0051) & \textbf{0.2113} (0.0054) & 2333.4 & 2228.3 & 2228.3 \\
70, 10, 5 & 0.2152 (0.0062) & 0.2121 (0.0065) & \textbf{0.2107} (0.0063) & 2373.4 & 2260.3 & 2260.3 \\
80, 10, 5 & 0.2129 (0.0080) & 0.2121 (0.0037) & \textbf{0.2113} (0.0049) & 2413.4 & 2292.3 & 2292.3 \\
90, 10, 5 & 0.2118 (0.0055) & 0.2153 (0.0054) & \textbf{0.2103} (0.0059) & 2453.4 & 2324.3 & 2324.3 \\
100, 10, 5 & 0.2130 (0.0063) & \textbf{0.2101} (0.0034) & 0.2115 (0.0054) & 2493.4 & 2356.3 & 2356.3 \\
\bottomrule
\end{tabular}}
\end{table}

\begin{table}[H]
\centering
\caption{Repeated 10-fold cross-validation results for the \texttt{bike sharing} dataset.}
\label{tab:real-bike_sharing}
\resizebox{\ifdim\width>\textwidth\textwidth\else\width\fi}{!}{%
\begin{tabular}{lcccccc}
\toprule
 & \multicolumn{3}{c}{Avg. MSE (SD)} & \multicolumn{3}{c}{Avg. total number of bases} \\
\cmidrule(lr){2-4}\cmidrule(lr){5-7}
Number of knots & Quantile & $k$-means & PAM & Quantile & $k$-means & PAM \\
\midrule
20, 10, 5 & 428656 (15380) & \textbf{416109} (16174) & 427174 (11184) & 8387.5 & 7459.7 & 7459.7 \\
30, 10, 5 & 430422 (18482) & \textbf{415528} (13373) & 426037 (10391) & 8427.5 & 7479.7 & 7479.7 \\
40, 10, 5 & 426395 (18286) & \textbf{421613} (14871) & 425259 (10476) & 8467.5 & 7499.7 & 7499.7 \\
50, 10, 5 & 420270 (15761) & \textbf{416930} (14592) & 426131 (14272) & 8507.5 & 7519.7 & 7519.7 \\
60, 10, 5 & 418502 (13593) & \textbf{415577} (15114) & 428546 (13129) & 8547.5 & 7539.7 & 7539.7 \\
70, 10, 5 & 419519 (14943) & \textbf{416526} (15837) & 429209 (12011) & 8587.5 & 7559.7 & 7559.7 \\
80, 10, 5 & 415991 (12950) & \textbf{412263} (9918) & 427585 (12743) & 8627.5 & 7579.7 & 7579.7 \\
90, 10, 5 & 419276 (13439) & \textbf{415075} (13411) & 428498 (12088) & 8667.5 & 7599.7 & 7599.7 \\
100, 10, 5 & 418833 (16393) & \textbf{411326} (13626) & 423821 (11500) & 8707.5 & 7619.7 & 7619.7 \\
\bottomrule
\end{tabular}}
\end{table}

\begin{table}[H]
\centering
\caption{Repeated 10-fold cross-validation results for the \texttt{liver disorders} dataset.}
\label{tab:real-liver_disorders}
\resizebox{\ifdim\width>\textwidth\textwidth\else\width\fi}{!}{%
\begin{tabular}{lcccccc}
\toprule
 & \multicolumn{3}{c}{Avg. MSE (SD)} & \multicolumn{3}{c}{Avg. total number of bases} \\
\cmidrule(lr){2-4}\cmidrule(lr){5-7}
Number of knots & Quantile & $k$-means & PAM & Quantile & $k$-means & PAM \\
\midrule
20, 10, 5 & \textbf{8.893} (0.144) & 8.986 (0.172) & 8.982 (0.131) & 1701.3 & 1689.2 & 1689.2 \\
30, 10, 5 & \textbf{8.895} (0.141) & 9.011 (0.165) & 9.004 (0.145) & 1742.8 & 1730.7 & 1730.7 \\
40, 10, 5 & \textbf{8.938} (0.131) & 8.950 (0.104) & 8.978 (0.215) & 1764.9 & 1748.5 & 1748.5 \\
50, 10, 5 & \textbf{8.876} (0.132) & 9.012 (0.147) & 8.984 (0.138) & 1800.6 & 1777.0 & 1777.0 \\
60, 10, 5 & \textbf{8.893} (0.117) & 9.024 (0.131) & 8.989 (0.191) & 1812.7 & 1786.8 & 1786.8 \\
70, 10, 5 & \textbf{8.892} (0.118) & 8.940 (0.124) & 8.995 (0.154) & 1845.6 & 1811.3 & 1811.3 \\
80, 10, 5 & \textbf{8.912} (0.113) & 9.069 (0.130) & 8.980 (0.188) & 1873.8 & 1828.4 & 1828.4 \\
90, 10, 5 & \textbf{8.885} (0.130) & 9.027 (0.261) & 8.990 (0.150) & 1892.4 & 1839.5 & 1839.5 \\
100, 10, 5 & \textbf{8.913} (0.139) & 9.001 (0.161) & 8.983 (0.147) & 1910.2 & 1850.1 & 1850.1 \\
\bottomrule
\end{tabular}}
\end{table}

\begin{table}[H]
\centering
\caption{Repeated 10-fold cross-validation results for the \texttt{laheart} dataset.}
\label{tab:real-laheart}
\resizebox{\ifdim\width>\textwidth\textwidth\else\width\fi}{!}{%
\begin{tabular}{lcccccc}
\toprule
 & \multicolumn{3}{c}{Avg. MSE (SD)} & \multicolumn{3}{c}{Avg. total number of bases} \\
\cmidrule(lr){2-4}\cmidrule(lr){5-7}
Number of knots & Quantile & $k$-means & PAM & Quantile & $k$-means & PAM \\
\midrule
20, 10, 5 & 0.1948 (0.0058) & 0.1945 (0.0039) & \textbf{0.1939} (0.0055) & 14412.1 & 14283.7 & 14283.7 \\
30, 10, 5 & 0.1944 (0.0059) & 0.1994 (0.0058) & \textbf{0.1924} (0.0051) & 14483.2 & 14339.6 & 14339.6 \\
40, 10, 5 & 0.1967 (0.0067) & 0.1995 (0.0060) & \textbf{0.1947} (0.0082) & 14561.7 & 14403.4 & 14403.4 \\
50, 10, 5 & 0.1946 (0.0044) & 0.1983 (0.0075) & \textbf{0.1922} (0.0057) & 14607.5 & 14436.1 & 14436.2 \\
60, 10, 5 & \textbf{0.1936} (0.0033) & 0.1984 (0.0060) & 0.1944 (0.0086) & 14658.4 & 14466.9 & 14466.9 \\
70, 10, 5 & \textbf{0.1947} (0.0044) & 0.1980 (0.0058) & 0.1955 (0.0076) & 14679.8 & 14478.5 & 14478.5 \\
80, 10, 5 & 0.1948 (0.0051) & 0.1949 (0.0054) & \textbf{0.1946} (0.0061) & 14702.6 & 14491.1 & 14491.2 \\
90, 10, 5 & 0.1958 (0.0067) & 0.1964 (0.0052) & \textbf{0.1957} (0.0087) & 14788.1 & 14532.0 & 14532.0 \\
100, 10, 5 & 0.1943 (0.0052) & 0.1975 (0.0046) & \textbf{0.1941} (0.0067) & 14796.5 & 14535.8 & 14535.9 \\
\bottomrule
\end{tabular}}
\end{table}

\begin{table}[H]
\centering
\caption{Repeated 10-fold cross-validation results for the \texttt{real estate} dataset.}
\label{tab:real-real_estate}
\resizebox{\ifdim\width>\textwidth\textwidth\else\width\fi}{!}{%
\begin{tabular}{lcccccc}
\toprule
 & \multicolumn{3}{c}{Avg. MSE (SD)} & \multicolumn{3}{c}{Avg. total number of bases} \\
\cmidrule(lr){2-4}\cmidrule(lr){5-7}
Number of knots & Quantile & $k$-means & PAM & Quantile & $k$-means & PAM \\
\midrule
20, 10, 5 & 59.53 (1.43) & 59.86 (0.90) & \textbf{57.68} (0.60) & 2503.6 & 2475.5 & 2475.5 \\
30, 10, 5 & 58.89 (0.98) & 60.34 (1.40) & \textbf{57.17} (0.49) & 2542.5 & 2499.5 & 2499.5 \\
40, 10, 5 & 58.63 (0.72) & 59.83 (1.03) & \textbf{57.66} (0.68) & 2581.2 & 2527.0 & 2527.0 \\
50, 10, 5 & 58.51 (1.26) & 59.98 (1.09) & \textbf{57.42} (0.55) & 2618.4 & 2551.4 & 2551.4 \\
60, 10, 5 & 58.05 (1.43) & 59.82 (0.79) & \textbf{57.57} (0.54) & 2653.6 & 2575.0 & 2575.0 \\
70, 10, 5 & 57.98 (1.27) & 59.37 (1.29) & \textbf{57.44} (0.75) & 2687.1 & 2596.9 & 2596.9 \\
80, 10, 5 & 58.09 (0.85) & 59.91 (0.94) & \textbf{57.33} (0.74) & 2720.4 & 2619.1 & 2619.1 \\
90, 10, 5 & 57.86 (0.52) & 59.66 (1.16) & \textbf{57.32} (0.55) & 2752.5 & 2640.5 & 2640.5 \\
100, 10, 5 & 57.91 (0.87) & 59.20 (0.84) & \textbf{57.58} (0.95) & 2782.9 & 2660.7 & 2660.7 \\
\bottomrule
\end{tabular}}
\end{table}

\end{document}